\documentclass[runningheads]{llncs}

\usepackage{graphicx}
\usepackage{subcaption}
\usepackage{float}
\usepackage[justification=raggedright]{caption}	
\usepackage{lscape}                                         

\usepackage[lined,ruled,linesnumbered]{algorithm2e}

\usepackage{booktabs}                   
\usepackage{multirow}

\usepackage{paralist}
\usepackage{enumitem}

\usepackage{bm}                          
\usepackage{epsfig}                      
\usepackage{graphicx}                  
\usepackage{mathtools}
\usepackage{amssymb,amsmath}   

\usepackage{units}
\usepackage{color}

\usepackage{comment}

\usepackage{url}  
\usepackage[pagebackref=true,breaklinks=true,letterpaper=true,colorlinks,bookmarks=false]{hyperref}

\usepackage{xspace}
\usepackage[table]{xcolor}
\usepackage{setspace}

\def\etal{et~al.}			  
\def\eg{e.g.,~}               
\def\ie{i.e.,~}               

\newlength\paramargin
\newlength\figmargin
\newlength\secmargin
\newlength\figcapmargin

\newcommand{\secref}[1]{Section~\ref{sec:#1}}
\newcommand{\figref}[1]{Figure~\ref{fig:#1}}

\long\def\ignorethis#1{}

\newcommand{\tb}[1]{\textbf{#1}}

\renewcommand{\paragraph}{\vspace{2mm} \textbf}

\def\xi{\mathbf{x}_i}

\graphicspath{{figure}, {example}}

\usepackage{graphicx}
\usepackage{amsmath,amssymb} 
\usepackage{color}

\usepackage{listings}
 \definecolor{codegreen}{rgb}{0,0.6,0}
\definecolor{codegray}{rgb}{0.5,0.5,0.5}
\definecolor{codepurple}{rgb}{0.58,0,0.82}
\definecolor{backcolour}{rgb}{0.95,0.95,0.92}
 \lstdefinestyle{mystyle}{
    backgroundcolor=\color{backcolour},   
    commentstyle=\color{codegreen},
    keywordstyle=\color{magenta},
    numberstyle=\tiny\color{codegray},
    stringstyle=\color{codepurple},
    basicstyle=\footnotesize,
    breakatwhitespace=false,         
    breaklines=true,                 
    captionpos=b,                    
    keepspaces=true,                 
    numbers=left,                    
    numbersep=5pt,                  
    showspaces=false,                
    showstringspaces=false,
    showtabs=false,                  
    tabsize=2
}
\usepackage[width=122mm,left=12mm,paperwidth=146mm,height=193mm,top=12mm,paperheight=217mm]{geometry}
\usepackage{animate} 

\begin{document}


\newcommand{\jb}[1]{\textcolor{green}{JB: #1}}
\newcommand{\yl}[1]{\textcolor{red}{YL: #1}}
\newcommand{\para}[1]{\paragraph{#1}}
\newcommand{\red}{\textcolor{red}}
\newcommand{\blue}{\textcolor{blue}}
\newcommand{\orange}{\textcolor{orange}}

\title{DF-Net: Unsupervised Joint Learning of \\Depth and Flow using Cross-Task Consistency} 
\titlerunning{Unsupervised Joint Learning using Cross-Task Consistency}

\author{
Yuliang Zou\inst{1}
\and 
Zelun Luo\inst{2}
\and
Jia-Bin Huang\inst{1}
}
\authorrunning{Y. Zou, Z. Luo, and J.-B. Huang}

\institute{Virginia Tech \and Stanford University}
\institute{$^1$Virginia Tech~\quad~$^2$Stanford University}

\newcommand{\mpage}[2]
{
\begin{minipage}{#1\linewidth}\centering
#2
\end{minipage}
}

\maketitle
\begin{center}
\centering
\mpage{0.01}{\rotatebox[origin=c]{90}{Depth}}
\mpage{0.3}{\includegraphics[width=\linewidth]{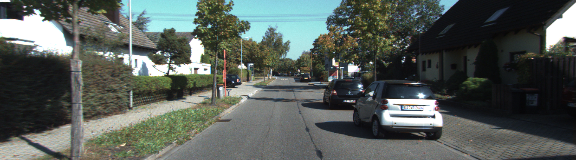}} 
\hfill
\mpage{0.3}{\includegraphics[width=\linewidth]{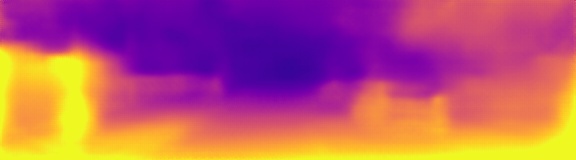}} 
\hfill
\mpage{0.3}{\includegraphics[width=\linewidth]{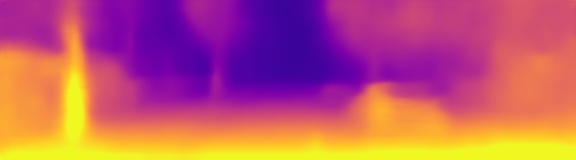}}
\\

\mpage{0.01}{\rotatebox[origin=c]{90}{Flow}}
\mpage{0.3}
{\includegraphics[width=\linewidth]{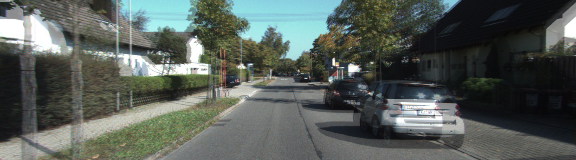}} 
\hfill
\mpage{0.3}{\includegraphics[width=\linewidth]{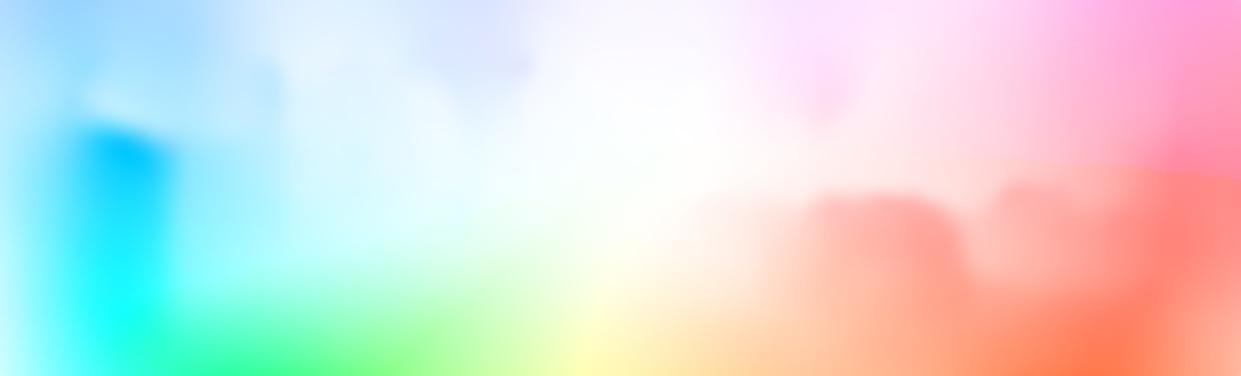}} 
\hfill
\mpage{0.3}{\includegraphics[width=\linewidth]{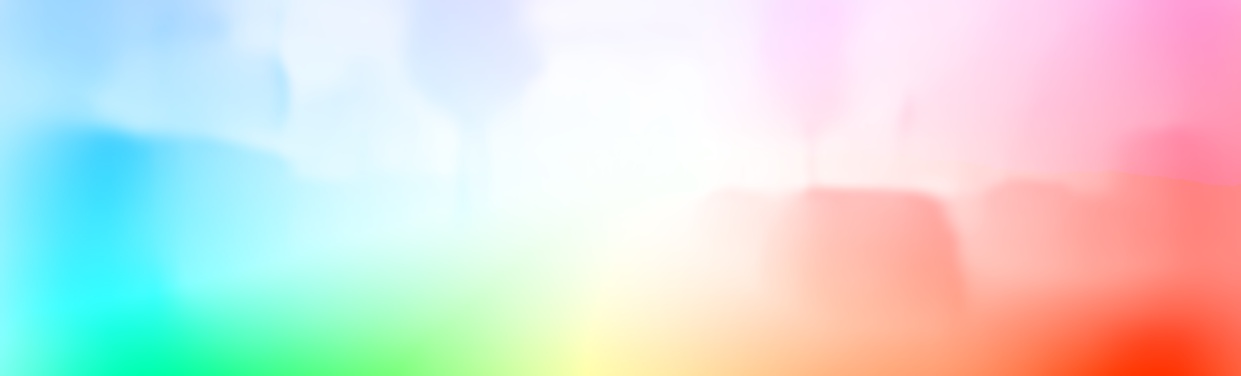}}
\\
\mpage{0.01}{\rotatebox[origin=c]{90}{}}
\mpage{0.3}{Input} 
\hfill
\mpage{0.3}{Separate learning} 
\hfill
\mpage{0.3}{Joint learning (Ours)}\\

\captionof{figure}{\tb{Joint learning v.s. separate learning.}
Single-view depth prediction and optical flow estimation are two highly correlated tasks.
Existing work, however, often addresses these two tasks in isolation.
In this paper, we propose a novel cross-task consistency loss to couple the training of these two problems using unlabeled monocular videos.
Through enforcing the underlying geometric constraints, we show substantially improved results for both tasks. 
}

\label{fig:teaser}
\end{center}

\begin{abstract}

We present an unsupervised learning framework for simultaneously training single-view depth prediction and optical flow estimation models using unlabeled video sequences.
Existing unsupervised methods often exploit brightness constancy and spatial smoothness priors to train depth or flow models.
In this paper, we propose to leverage geometric consistency as additional supervisory signals.
Our core idea is that for rigid regions we can use the predicted scene depth and camera motion to synthesize 2D optical flow by backprojecting the induced 3D scene flow.
The discrepancy between the rigid flow (from depth prediction and camera motion) and the estimated flow (from optical flow model) allows us to impose a cross-task consistency loss.
While all the networks are jointly optimized during training, they can be applied independently at test time.
%
Extensive experiments demonstrate that our depth and flow models compare favorably with state-of-the-art unsupervised methods.

\end{abstract}

\section{Introduction}
\label{sec:intro}


Single-view depth prediction and optical flow estimation are two fundamental problems in computer vision.
While the two tasks aim to recover highly correlated information from the scene (\ie the scene structure and the dense motion field between consecutive frames), existing efforts typically study each problem in isolation.
%
In this paper, we demonstrate the benefits of exploring the geometric relationship between depth, camera motion, and flow for unsupervised learning of depth and flow estimation models.
%
%

\begin{figure}[t]
\centering
\mpage{0.45}{\includegraphics[width=0.8\linewidth]{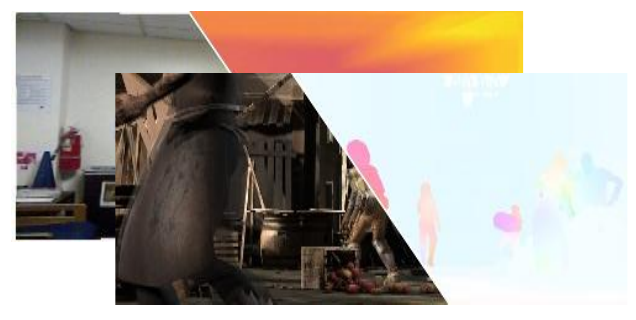}}\hfill
\mpage{0.45}{\includegraphics[width=\linewidth]{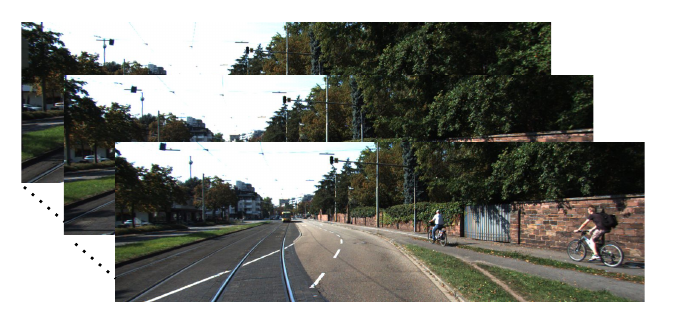}}

\mpage{0.45}{Pixelwise ground truth} 
\hfill
\mpage{0.45}{Unlabeled video sequences (ours)}
\caption{\textbf{Supervised v.s. unsupervised learning.}
Supervised learning of depth or flow networks requires large amount of training data with pixelwise ground truth annotations, which are difficult to acquire in real scenes. 
%
%
In contrast, our work leverages the readily available \emph{unlabeled} video sequences to jointly train the depth and flow models. 
}
\label{fig:motivation}
\end{figure}
With the rapid development of deep convolutional neural networks (CNNs), numerous approaches have been proposed to tackle dense prediction problems in an end-to-end manner.
However, supervised training CNN for such tasks often involves in constructing large-scale, diverse datasets with dense pixelwise ground truth labels.
Collecting such densely labeled datasets in real-world requires significant amounts of human efforts and is prone to error. 
Existing efforts of RGB-D dataset construction~\cite{Geiger2013IJRR,Silberman:ECCV12,saxena2006learning,saxena20083} often have limited scope (\eg in terms of locations, scenes, and objects), and hence are lack of diversity.
For optical flow, dense motion annotations are even more difficult to acquire~\cite{liu2008human}.
Consequently, existing CNN-based methods rely on synthetic datasets for training the models~\cite{Butler:ECCV:2012,DFIB15,Gaidon:Virtual:CVPR2016,huang2018deepmvs}.
These synthetic datasets, however, do not capture the complexity of motion blur, occlusion, and natural image statistics from real scenes.
The trained models usually do not generalize well to unseen scenes without fine-tuning on sufficient ground truth data in a new visual domain.

Several work~\cite{garg2016unsupervised,godard2016unsupervised,jason2016back} have been proposed to capitalize on large-scale real-world videos to train the CNNs in the unsupervised setting. 
The main idea lies to exploit the brightness constancy and spatial smoothness assumptions of flow fields or disparity maps as supervisory signals.
These assumptions, however, often do not hold at motion boundaries and hence makes the training unstable.

%
Many recent efforts~\cite{tung2017adversarial,vijayanarasimhan2017sfm,wulff2017optical,zhou2017unsupervised} explore the geometric relationship between the two problems.
With the estimated depth and camera pose, these methods can produce dense optical flow by backprojecting the 3D scene flow induced from camera ego-motion.
However, these methods implicitly assume \emph{perfect} depth and camera pose estimation when ``synthesizing'' the optical flow.
The errors in either depth or camera pose estimation inevitably produce inaccurate flow predictions.

In this paper, we present a technique for \emph{jointly} learning a single-view depth estimation model and a flow prediction model using unlabeled videos as shown in \figref{motivation}.
Our key observation is that the predictions from depth, pose, and optical flow should be \emph{consistent} with each other.
By exploiting this geometry cue, we present a novel cross-task consistency loss that provides additional supervisory signals for training both networks.
We validate the effectiveness of the proposed approach through extensive experiments on several benchmark datasets.
Experimental results show that our joint training method significantly improves the performance of both models (\figref{teaser}).
The proposed depth and flow models compare favorably with state-of-the-art unsupervised methods.
%
%

We make the following contributions. (1) We propose an unsupervised learning framework to \emph{simultaneously} train a depth prediction network and an optical flow network. We achieve this by introducing a cross-task consistency loss that enforces geometric consistency. (2) We show that through the proposed unsupervised training our depth and flow models compare favorably with existing unsupervised algorithms and achieve competitive performance with supervised methods on several benchmark datasets. (3) We release the source code and pre-trained models to facilitate future research: \url{http://yuliang.vision/DF-Net/}

\section{Related Work}
\label{sec:related}

\paragraph{Supervised learning of depth and flow.} 
Supervised learning using CNNs has emerged to be an effective approach for depth and flow estimation to avoid hand-crafted objective functions and computationally expensive optimization at test time.
The availability of RGB-D datasets and deep learning leads to a line of work on single-view depth estimation~\cite{eigen2015predicting,eigen2014depth,li2015depth,liu2015deep,wang2015towards,zhang2015monocular}.
While promising results have been shown, these methods rely on the \emph{absolute} ground truth depth maps. 
These depth maps, however, are expensive and difficult to collect.
Some efforts~\cite{chen2016single,zoran2015learning} have been made to relax the difficulty of collecting absolute depth by exploring learning from \emph{relative/ordinal} depth annotations.
Recent work also explores gathering training datasets from web videos~\cite{chen2018learning} or Internet photos~\cite{li2018megadepth} using structure-from-motion and multi-view stereo algorithms.

Compared to ground truth depth datasets, constructing optical flow datasets of diverse scenes in real-world is even more challenging.
Consequently, existing approaches~\cite{DFIB15,ilg2016flownet,ranjan2016optical} typically rely on synthetic datasets~\cite{Butler:ECCV:2012,DFIB15} for training.
Due to the limited scalability of constructing diverse, high-quality training data, fully supervised approaches often require fine-tuning on sufficient ground truth labels in new visual domains to perform well.
In contrast, our approach leverages the readily available real-world videos to jointly train the depth and flow models.
The ability to learn from unlabeled data enables unsupervised pre-training for domains with limited amounts of ground truth data.
%

\paragraph{Self-supervised learning of depth and flow.}
To alleviate the dependency on large-scale annotated datasets, several works have been proposed to exploit the classical assumptions of brightness constancy and spatial smoothness on the disparity map or the flow field~\cite{garg2016unsupervised,godard2016unsupervised,jason2016back,meister2017unflow,zhan2018unsupervised}.
The core idea is to treat the estimated depth and flow as latent layers and use them to differentiably warp the source frame to the target frame, where the source and target frames can either be the stereo pair or two consecutive frames in a video sequence.
A photometric loss between the synthesized frame and the target frame can then serve as an unsupervised proxy loss to train the network.
Using photometric loss alone, however, is not sufficient due to the ambiguity on textureless regions and occlusion boundaries.
Hence, the network training is often unstable and requires careful hyper-parameter tuning of the loss functions.
Our approach builds upon existing unsupervised losses for training our depth and flow networks.
We show that the proposed cross-task consistency loss provides a sizable performance boost over individually trained models.

\paragraph{Methods exploiting geometry cues.}
Recently, a number of work exploits the geometric relationship between depth, camera pose, and flow for learning depth or flow models~\cite{vijayanarasimhan2017sfm,wulff2017optical,yin2018geonet,zhou2017unsupervised}.
These methods first estimate the depth of the input images.
Together with the estimated camera poses between two consecutive frames, these methods ``synthesize'' the flow field of rigid regions.
The synthesized flow from depth and pose can either be used for flow prediction in rigid regions~\cite{vijayanarasimhan2017sfm,wulff2017optical,yin2018geonet,ranjan2018adversarial} as is or used for view synthesis to train depth model using monocular videos~\cite{zhou2017unsupervised}.
%
%
Additional cues such as surface normal~\cite{yang2017unsupervised},  edge~\cite{yang2018lego}, physical constraints~\cite{tung2017adversarial} can be incorporated to further improve the performance.
%

These approaches exploit the inherent geometric relationship between structure and motion. 
However, the errors produced by either the depth or the camera pose estimation propagate to flow predictions. 
Our key insight is that for rigid regions the estimated flow (from flow prediction network) and the synthesized rigid flow (from depth and camera pose networks) should be consistent.
Consequently, coupled training allows both depth and flow networks to learn from each other and enforce geometrically consistent predictions of the scene.
%

\paragraph{Structure from motion.} 
Joint estimation of structure and camera pose from multiple images of a given scene is a long-standing problem~\cite{newcombe2011dtam,furukawa2010towards,wu2011visualsfm}.
Conventional methods can recover (semi-)dense depth estimation and camera pose through keypoint tracking/matching.
The outputs of these algorithms can potentially be used to help train a flow network, but not the other way around. 
Our work differs as we are also interested in learning a depth network to recover dense structure from a single input image.

%

\paragraph{Multi-task learning.} 
Simultaneously addressing multiple tasks through multi-task learning~\cite{ruder2017overview} has shown advantages over methods that tackle individual ones~\cite{zamir2018taskonomy}.
For examples, joint learning of video segmentation and optical flow through layered models~\cite{chang2013topology,sun2013fully} or feature sharing~\cite{chengsegflow} helps improve accuracy at motion boundaries. 
Single-view depth model learning can also benefit from joint training with surface normal estimation~\cite{li2015depth,yang2017unsupervised} or semantic segmentation~\cite{eigen2015predicting,kendall2017multi}.

Our approach tackles the problems of learning both depth and flow models.
Unlike existing multi-task learning methods that often require \emph{direct supervision} using ground truth training data for each task, our approach instead leverage \emph{meta-supervision} to couple the training of depth and flow models.
While our models are jointly trained, they can be applied independently at test time.

\section{Unsupervised Joint Learning of Depth and Flow}
\label{sec:method}

\subsection{Method overview}
\label{sec:overview}

\begin{figure*}[t]
\centering
\includegraphics[width=0.94\linewidth]{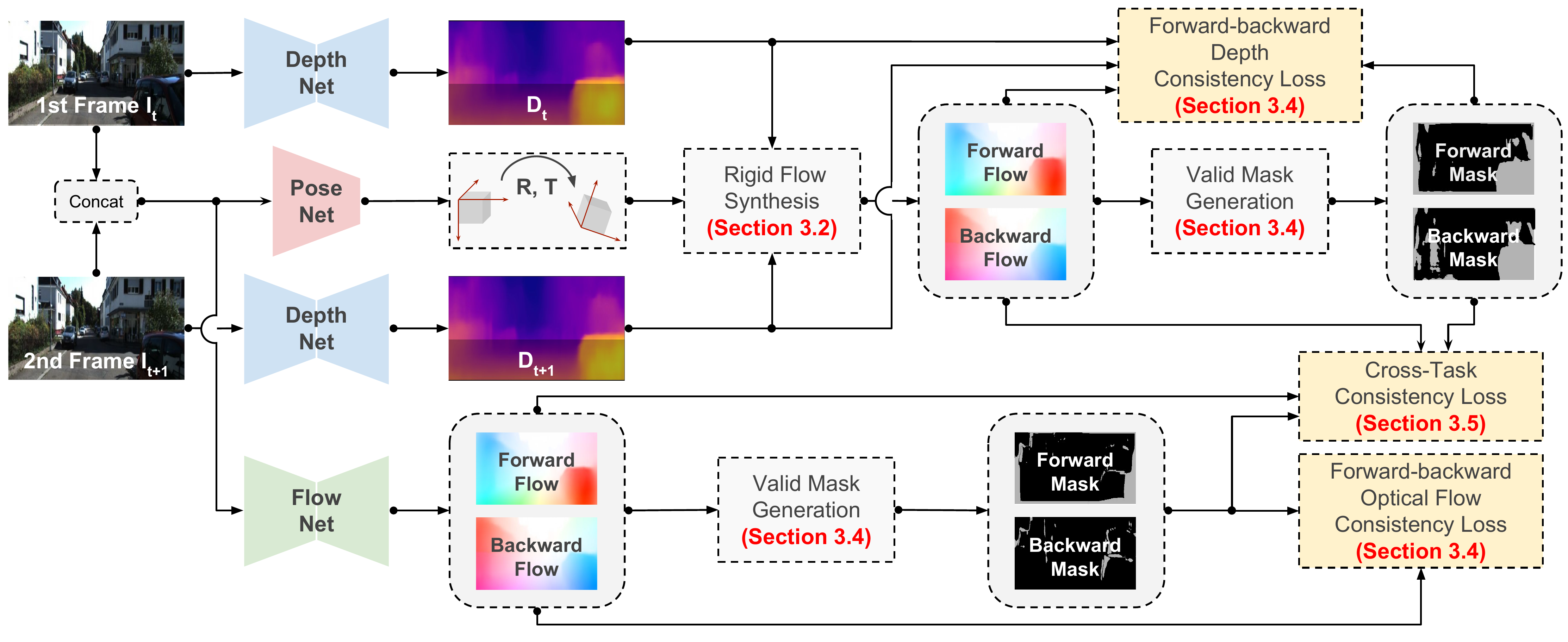}
\caption{\textbf{Overview of our unsupervised joint learning framework.}
Our framework consists of three major modules: 
(1) a \emph{Depth Net} for single-view depth estimation; 
(2) a \emph{Pose Net} that takes two stacked input frames and estimates the relative camera pose between the two input frames; and 
(3) a \emph{Flow Net} that estimates dense optical flow field between the two input frames. 
Given a pair of input images $\mathbf{I}_t$ and $\mathbf{I}_{t+1}$ sampled from an unlabeled video, we first estimate the depth of each frame, the 6D camera pose, and the dense forward and backward flows.
Using the predicted scene depth and the estimated camera pose, we can synthesize 2D forward and backward optical flows (referred as \emph{rigid flow}) by backprojecting the induced 3D forward and backward scene flows (\secref{synthesis}).
As we do not have ground truth depth and flow maps for supervision, we leverage standard photometric and spatial smoothness costs to regularize the network training (\secref{photometric}, not shown in this figure for clarity).
To enforce the consistency of flow and depth prediction in both directions, we exploit the forward-backward consistency (\secref{forward}), and adopt the valid masks derived from it to filter out invalid regions (e.g., occlusion/dis-occlusion) for the photometric loss.
Finally, we propose a novel cross-network consistency loss (\secref{consistency}) --- encouraging the optical flow estimation (from the \emph{Flow Net}) and the rigid flow (from the \emph{Depth and Pose Net}) to be consistent to each other within in valid regions.
}
\label{fig:model}
\end{figure*}

Our goal is to develop an unsupervised learning framework for \emph{jointly} training the single-view depth estimation network and the optical flow prediction network using \emph{unlabeled} video sequences.
\figref{model} shows the high-level sketch of our proposed approach.
Given two consecutive frames $(I_t, I_{t+1})$ sampled from an unlabeled video, we first estimate depth of frame $I_t$ and $I_{t+1}$, and forward-backward optical flow fields between frame $I_t$ and $I_{t+1}$.
We then estimate the 6D camera pose transformation between the two frames $(I_t, I_{t+1})$.

With the predicted depth map and the estimated 6D camera pose, we can produce the 3D scene flow induced from camera ego-motion and backproject them onto the image plane to synthesize the 2D flow (\secref{synthesis}).
We refer this synthesized flow as \emph{rigid flow}.
Suppose the scenes are mostly static, the synthesized rigid flow should be consistent with the results from the estimated optical flow (produced by the optical flow prediction model).
However, the prediction results from the two branches may not be consistent with each other.
%
Our intuition is that the discrepancy between the rigid flow and the estimated flow provides additional supervisory signals for both networks. 
Hence, we propose a \emph{cross-task consistency loss} to enforce this constraint (\secref{consistency}). 
To handle non-rigid transformations that cannot be explained by the camera motion and occlusion-disocclusion regions, we exploit the forward-backward consistency check to identify valid regions (\secref{forward}). 
We avoid enforcing the cross-task consistency for those forward-backward inconsistent regions.

Our overall objective function can be formulated as follows:
\begin{equation}
L = L_\text{photometric} + \lambda_{s} L_\text{smooth} + \lambda_{f}L_\text{forward-backward} + \lambda_{c} L_\text{cross}.
\end{equation}
All of the four loss terms are applied to both depth and flow networks. 
Also, all of the four loss terms are symmetric for forward and backward directions, for simplicity we only derive them for the forward direction.


\subsection{Flow synthesis using depth and pose predictions}
\label{sec:synthesis}
Given the two input frames $I_t$ and $I_{t+1}$, the predicted depth map $\hat{D}_t$, and relative camera pose $\hat{T}_{t\rightarrow t+1}$, here we wish to establish the dense pixel correspondence between the two frames.
Let $p_t$ denotes the 2D homogeneous coordinate of an pixel in frame $I_t$ and $K$ denotes the intrinsic camera matrix. 
We can compute the corresponding point of $p_t$ in frame $I_{t+1}$ using the equation~\cite{zhou2017unsupervised}:
\begin{equation}
p_{t+1} = K \hat{T}_{t\rightarrow t+1} \hat{D}_t(p_t) K^{-1} p_t.
\label{eqn:mapping}
\end{equation}

We can then obtain the synthesized forward rigid flow at pixel $p_t$ in $I_t$ by
\begin{equation}
F_\text{rigid}(p_t) = p_{t+1} - p_t
\end{equation}

\subsection{Brightness constancy and spatial smoothness priors}
\label{sec:photometric}
Here we briefly review two loss functions that we used in our framework to regularize network training.
Leveraging the brightness constancy and spatial smoothness priors used in classical dense correspondence algorithms~\cite{bruhn2005lucas,horn1981determining,lucas1981iterative}, prior work has used the photometric discrepancy between the warped frame and the target frame as an unsupervised proxy loss function for training CNNs without ground truth annotations. 

\paragraph{Photometric loss.}
Suppose that we have frame $I_t$ and $I_{t+1}$, as well as the estimated flow $F_{t\rightarrow t+1}$ (either from the optical flow predicted from the flow model or the synthesized rigid flow induced from the estimated depth and camera pose), we can produce the warped frame $\bar{I}_t$ with the inverse warping from frame $I_{t+1}$. 
%
%
Note that the projected image coordinates $p_{t+1}$ might not lie exactly on the image pixel grid, we thus apply a differentiable bilinear interpolation strategy used in the spatial transformer networks~\cite{jaderberg2015spatial} to perform frame synthesis. 

With the warped frame $\bar{I}_t$ from $I_{t+1}$, we formulate the brightness constancy objective function as
\begin{equation}
L_\text{photometric} = \sum_{p}\rho{\left(I_{t}(p), \bar{I}_{t}(p)\right)}.
\end{equation}
where $\rho(\cdot)$ is a function to measure the difference between pixel values. 
Previous work simply choose $L_1$ norm or the appearance matching loss~\cite{godard2016unsupervised}, which is not invariant to illumination changes in real-world scenarios~\cite{vogel2013evaluation}. 
Here we adopt the ternary census transform based loss~\cite{meister2017unflow,stein2004efficient,zabih1994non} that can better handle complex illumination changes.

\paragraph{Smoothness loss.}
The brightness constancy loss is not informative in low-texture or homogeneous region of the scene.
To handle this issue, existing work incorporates a smoothness prior to regularize the estimated disparity map or flow field.
We adopt the spatial smoothness loss as proposed in \cite{godard2016unsupervised}.
%
%
%

%


\subsection{Forward-backward consistency}
\label{sec:forward}
According to the brightness constancy assumption, the warped frame should be similar to the target frame.
However, the assumption does not hold for occluded and dis-occluded regions.
%
%
We address this problem by using the commonly used forward-backward consistency check technique to identify invalid regions and do not impose the photometric loss on those regions.

\paragraph{Valid masks.} We implement the occlusion detection based on forward-backward consistency assumption~\cite{sundaram2010dense} (i.e., traversing flow vector forward and then backward should arrive at the same position).
Here we use a simple criterion proposed in~\cite{meister2017unflow}.
%
We mark pixels as invalid whenever this constraint is violated. 
%
%
%
\figref{mask} shows two examples of the marked invalid regions by forward-backward consistency check using the synthesized rigid flow (animations can be viewed in Adobe Reader).


\begin{figure}[t]



\mpage{0.47}{\animategraphics[autoplay,loop,width=\textwidth]{6}{video/im_seq1/}{00030}{00036}}\hfill
\mpage{0.47}{\animategraphics[autoplay,loop,width=\textwidth]{6}{video/mask_seq1/}{00030}{00036}}\hfill

\mpage{0.47}{\animategraphics[autoplay,loop,width=\textwidth]{6}{video/im_seq2/}{00035}{00040}}\hfill
\mpage{0.47}{\animategraphics[autoplay,loop,width=\textwidth]{6}{video/mask_seq2/}{00035}{00040}}\hfill

\mpage{0.47}{Input frames} \hfill
\mpage{0.47}{Invalid masks by rigid flow}

\caption{\textbf{Valid mask visualization.}
We estimate the invalid mask by checking the forward-backward consistency from the synthesized rigid flow, which can not only detect occluded regions, but also identify the moving objects (cars) as they cannot be explained by the estimated depth and pose.
\blue{Animations can be viewed in Adobe Reader.}
%
}
\label{fig:mask}
\end{figure}

Denote the valid region by $V$ (either from rigid flow or estimated flow), we can modify the photometric loss term (4) as
\begin{equation}
L_\text{photometric} = \sum_{p\in V}\rho{\left(I_{t}(p), \bar{I}_{t}(p)\right)}.
\end{equation}


\paragraph{Forward-backward consistency loss.}
In addition to using forward-backward consistency check for identifying invalid regions, we can further impose constraints on the valid regions so that the network can produce consistent predictions for both forward and backward directions.
Similar ideas have been exploited in~\cite{hur2017mirrorflow,meister2017unflow} for occlusion-aware flow estimation. 
Here, we apply the forward-backward consistency loss to both flow and depth predictions.

For flow prediction, the forward-backward consistency loss is of the form:
\begin{equation}
L_\text{forward-backward, flow} =
\sum_{p\in V_\mathrm{flow}}\left||F_{t\rightarrow t+1}(p) + F_{t+1\rightarrow t}(p+F_{t\rightarrow t+1}(p))\right||_1
\end{equation}
Similarly, we impose a consistency penalty for depth:
\begin{equation}
L_\text{forward-backward, depth} =
\sum_{p\in V_\mathrm{depth}}||D_{t}(p) - \bar{D}_{t}(p)||_1
\end{equation}
where $\bar{D}_t$ is warped from $D_{t+1}$ using the synthesized rigid flow from $t$ to $t+1$.

While we exploit robust functions for enforcing photometric loss, forward-backward consistency for each of the tasks, the training of depth and flow networks using unlabeled data remains non-trivial and sensitive to the choice of hyper-parameters~\cite{Lai-NIPS-2017}.
Building upon the existing loss functions, in the following we introduce a novel cross-task consistency loss to further regularize the network training.

\subsection{Cross-task consistency}
\label{sec:consistency}
In \secref{synthesis}, we show that the motion of rigid regions in the scene can be explained by the ego-motion of the camera and the corresponding scene depth. 
On the one hand, we can estimate the rigid flow by backprojecting the induced 3D scene flow from the estimated depth and relative camera pose. 
On the other hand, we have direct estimation results from an optical flow network.
Our core idea is the that these two flow fields should be consistent with each other for non-occluded and static regions.
Minimizing the discrepancy between the two flow fields allows us to simultaneously update the depth and flow models. 

We thus propose to minimize the endpoint distance between the flow vectors in the rigid flow (computed from the estimated depth and pose) and that in the estimated flow (computed from the flow prediction model).
We denote the synthesized rigid flow as $F_\text{rigid}=(u_\text{rigid},v_\text{rigid})$ and the estimated flow as $F_\text{flow}=(u_\text{flow},v_\text{flow})$. 
Using the computed valid masks (\secref{forward}), we impose the cross-task consistency constraints over valid pixels.
\begin{equation}
L_\text{cross} = \sum_{p\in V_\mathrm{depth}\cap V_\mathrm{flow}}||F_\text{rigid}(p) - F_\text{flow}(p)||_1
\end{equation}

\section{Experimental Results}
\label{sec:results}

In this section, we validate the effectiveness of our proposed method for unsupervised learning of depth and flow on several standard benchmark datasets.
%
%
More results can be found in the supplementary material. 
Our source code and pre-trained models are available on \url{http://yuliang.vision/DF-Net/}.


\subsection{Datasets}
\paragraph{Datasets for joint network training.} 
We use video clips from the train split of  KITTI raw dataset~\cite{Geiger2013IJRR} for joint learning of depth and flow models. 
%
%
Note that our training does not involve any depth/flow labels.

\paragraph{Datasets for pre-training.} 
To avoid the joint training process converging to trivial solutions, we (unsupervisedly) pre-train the flow network on the SYNTHIA dataset~\cite{Ros_2016_CVPR}.
For pre-training both depth and pose networks, we use either KITTI raw dataset
or the CityScapes dataset~\cite{Cordts2016Cityscapes} 
.

%
The SYNTHIA dataset~\cite{Ros_2016_CVPR} contains multi-view frames captured by driving vehicles in different scenarios and traffic conditions.
We take all the four-view images of the left camera from all summer and winter driving sequences, which contains around 37K image pairs.
The CityScapes dataset~\cite{Cordts2016Cityscapes} contains real-world driving sequences, we follow Zhou~\etal~\cite{zhou2017unsupervised} and pre-process the dataset to generate around 75K training image pairs.

\paragraph{Datasets for evaluation.}
For evaluating the performance of our depth network, we use the \emph{test split} of the KITTI raw dataset. 
The depth maps for KITTI raw are sampled at irregularly spaced positions, captured using a rotating LIDAR scanner.
Following the standard evaluation protocol, we evaluate the performance using only the regions with ground truth depth samples (bottom parts of the images).
%
We also evaluate the generalization of our depth network on general scenes using the Make3D dataset~\cite{saxena2006learning}. 

For evaluating our flow network, we use the challenging KITTI flow 2012~\cite{Geiger2012CVPR} and KITTI flow 2015~\cite{Menze2015CVPR} datasets. 
The ground truth optical flow is obtained from a 3D laser scanner and thus only covers about 50\% of the pixels. 

\begin{figure*}[t]
\centering

\mpage{0.19}{\includegraphics[width=1.0\linewidth]{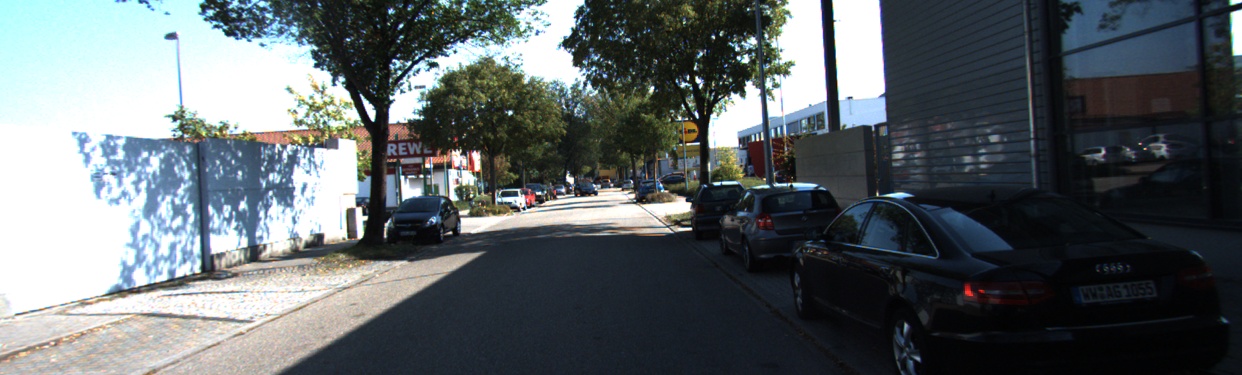}}\hfill
\mpage{0.19}{\includegraphics[width=1.0\linewidth]{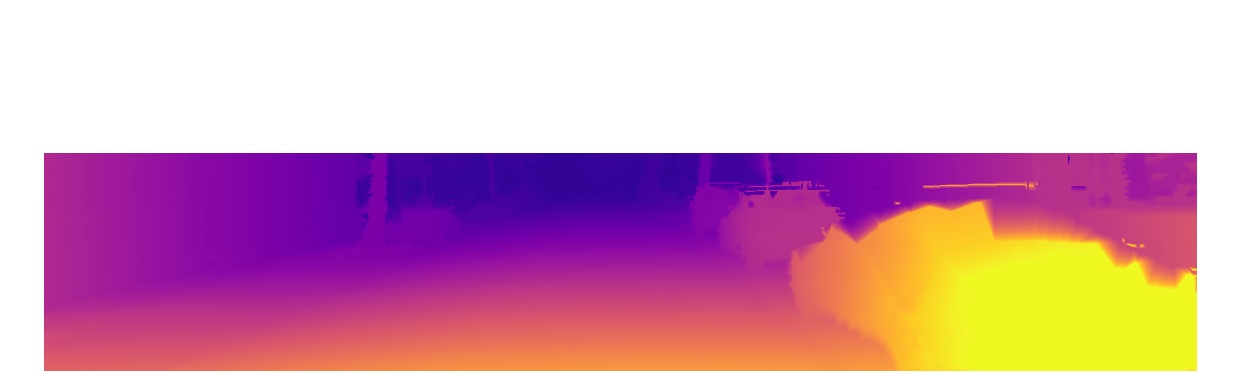}}\hfill
\mpage{0.18}{\includegraphics[width=1.0\linewidth]{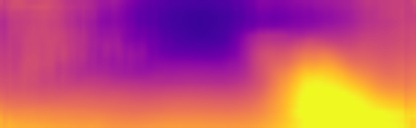}}\hfill
\mpage{0.18}{\includegraphics[width=1.0\linewidth]{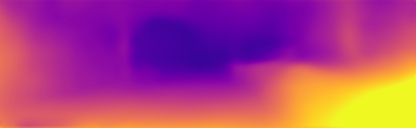}}\hfill
\mpage{0.18}{\includegraphics[width=1.0\linewidth]{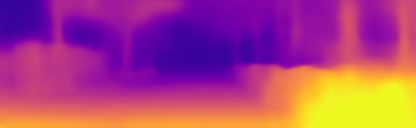}}\hfill


\mpage{0.19}{\includegraphics[width=1.0\linewidth]{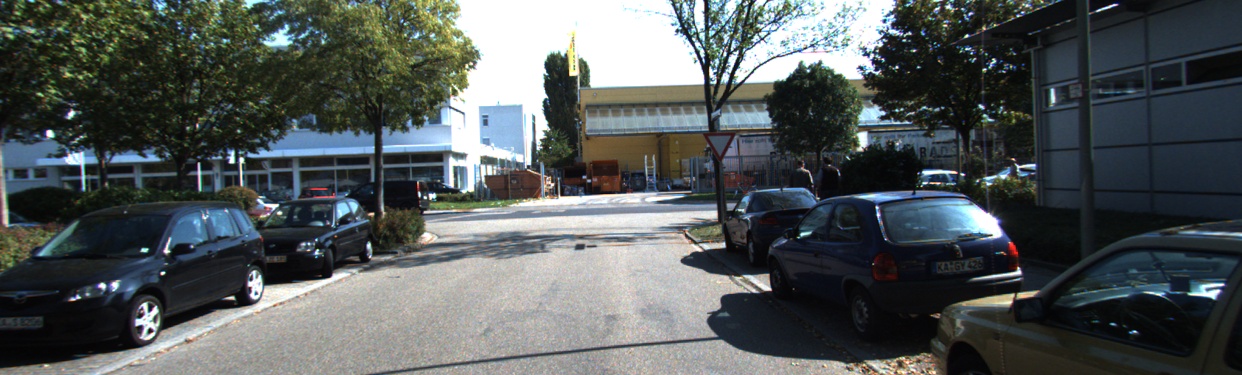}}\hfill
\mpage{0.19}{\includegraphics[width=1.0\linewidth]{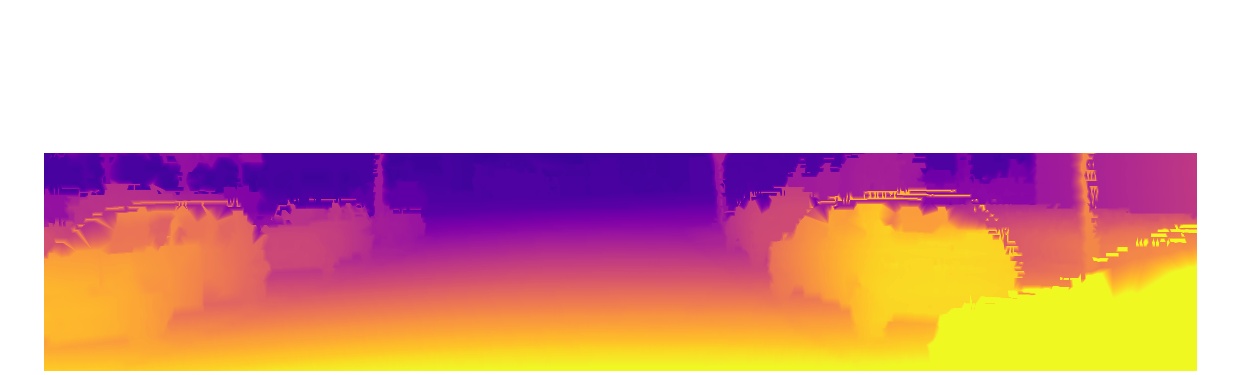}}\hfill
\mpage{0.18}{\includegraphics[width=1.0\linewidth]{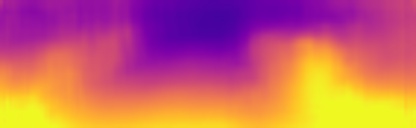}}\hfill
\mpage{0.18}{\includegraphics[width=1.0\linewidth]{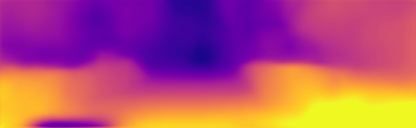}}\hfill
\mpage{0.18}{\includegraphics[width=1.0\linewidth]{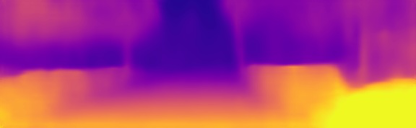}}\hfill



\mpage{0.19}{\includegraphics[width=1.0\linewidth]{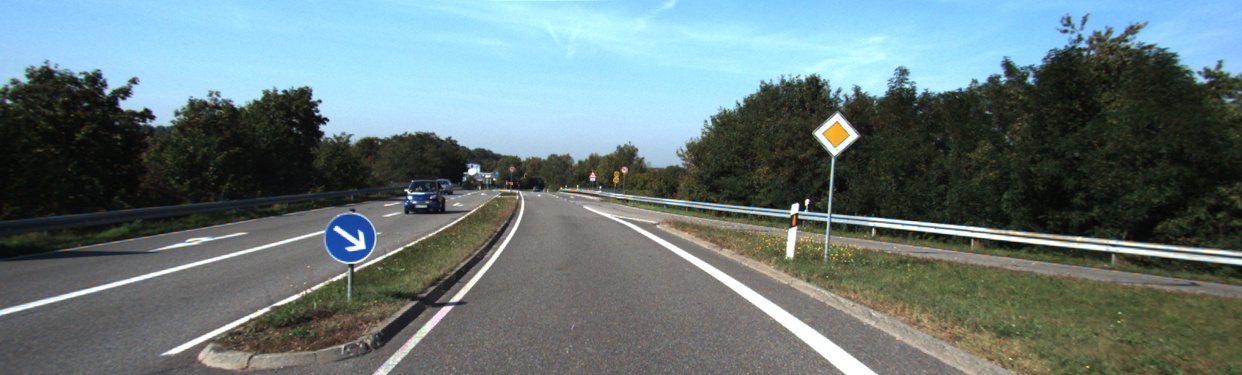}}\hfill
\mpage{0.19}{\includegraphics[width=1.0\linewidth]{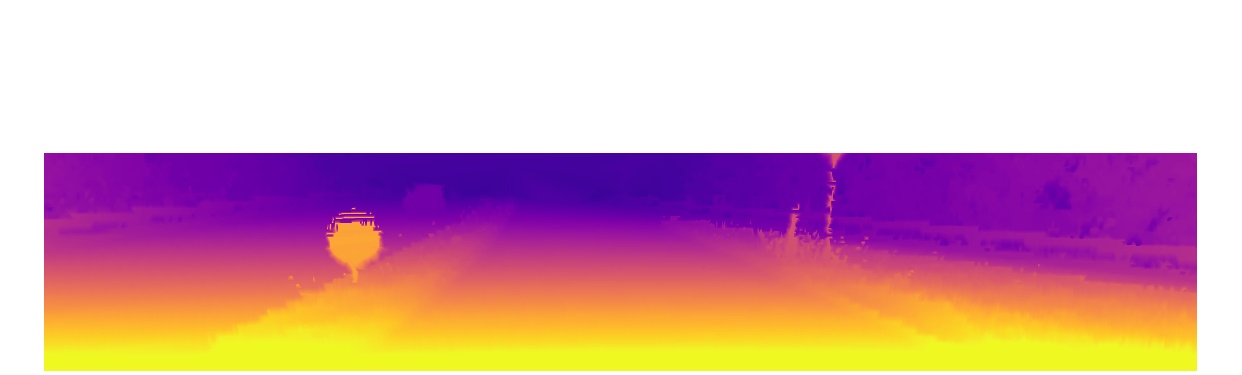}}\hfill
\mpage{0.18}{\includegraphics[width=1.0\linewidth]{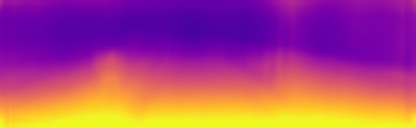}}\hfill
\mpage{0.18}{\includegraphics[width=1.0\linewidth]{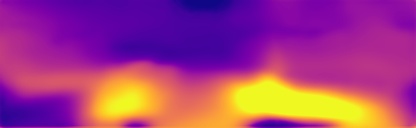}}\hfill
\mpage{0.18}{\includegraphics[width=1.0\linewidth]{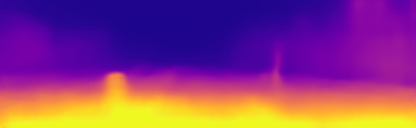}}\hfill


\mpage{0.19}{\includegraphics[width=1.0\linewidth]{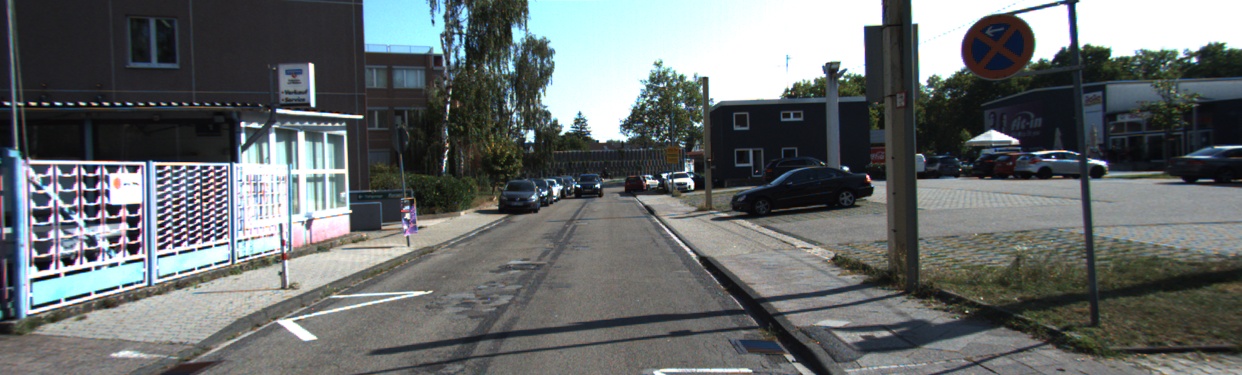}}\hfill
\mpage{0.19}{\includegraphics[width=1.0\linewidth]{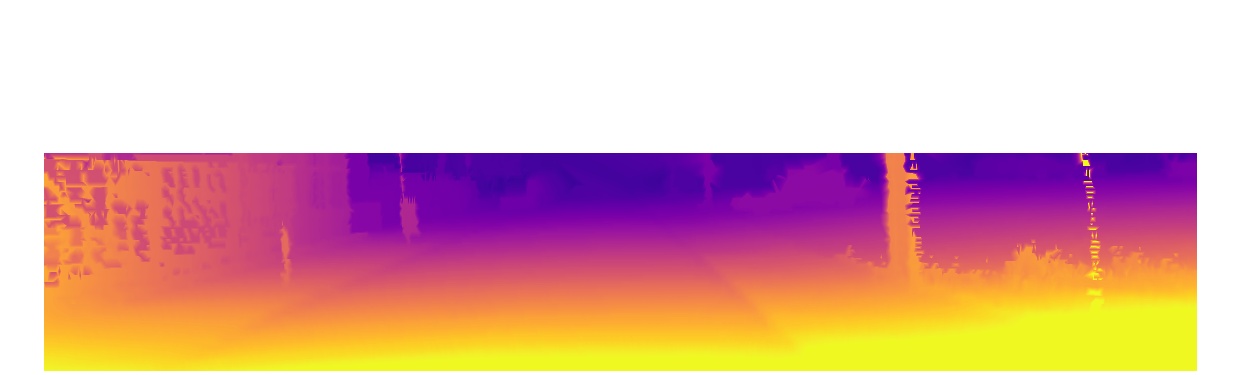}}\hfill
\mpage{0.18}{\includegraphics[width=1.0\linewidth]{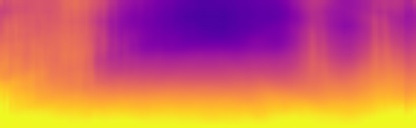}}\hfill
\mpage{0.18}{\includegraphics[width=1.0\linewidth]{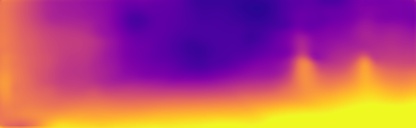}}\hfill
\mpage{0.18}{\includegraphics[width=1.0\linewidth]{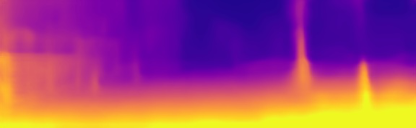}}\hfill


\mpage{0.19}{\includegraphics[width=1.0\linewidth]{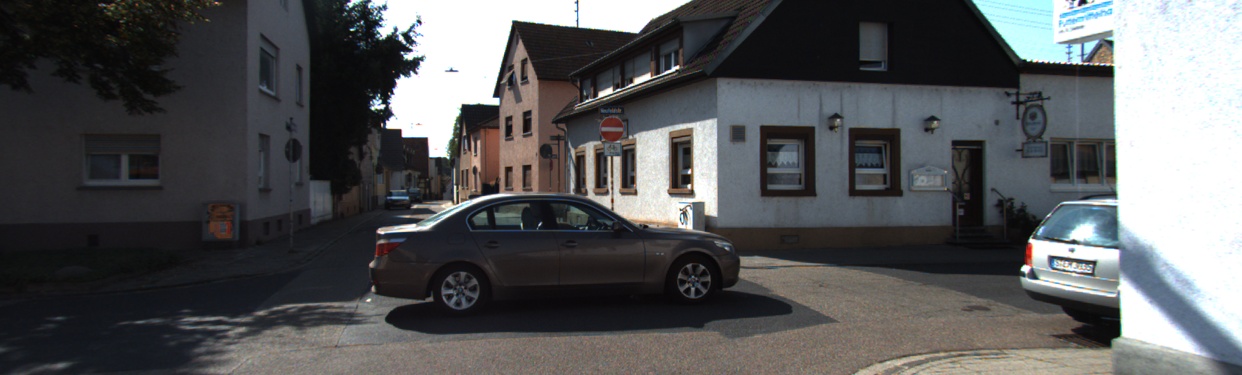}}\hfill
\mpage{0.19}{\includegraphics[width=1.0\linewidth]{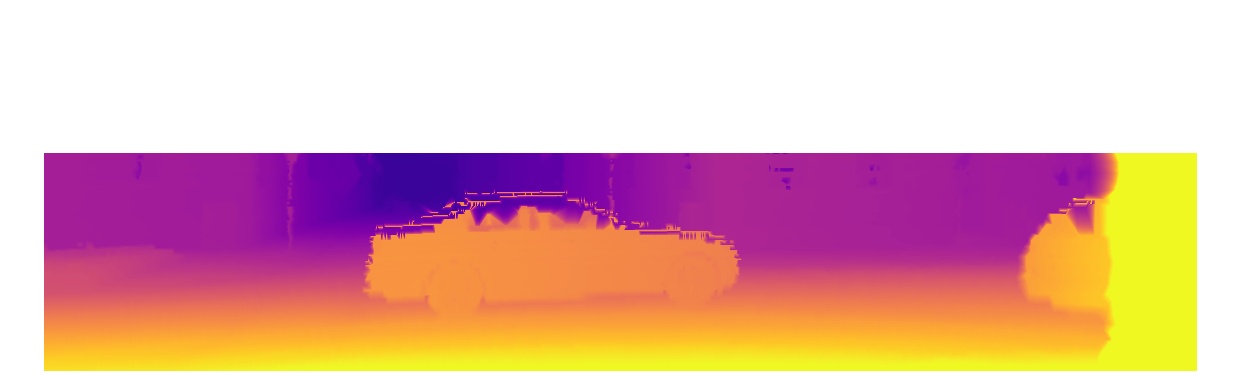}}\hfill
\mpage{0.18}{\includegraphics[width=1.0\linewidth]{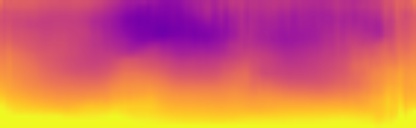}}\hfill
\mpage{0.18}{\includegraphics[width=1.0\linewidth]{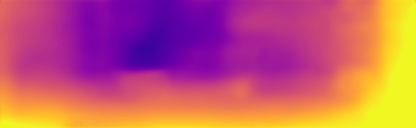}}\hfill
\mpage{0.18}{\includegraphics[width=1.0\linewidth]{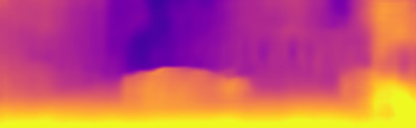}}\hfill

\mpage{0.18}{Input} \hfill
\mpage{0.18}{Ground truth} \hfill
\mpage{0.18}{Eigen~\etal~\cite{eigen2014depth}}\hfill
\mpage{0.19}{Zhou~\etal~\cite{zhou2017unsupervised}} \hfill
\mpage{0.16}{Ours}

\caption{\textbf{Sample results on KITTI raw test set.} The ground truth depth is interpolated from sparse point cloud for visualization only.
Compared to Zhou~\etal~\cite{zhou2017unsupervised} and Eigen~\etal~\cite{eigen2014depth}, our method can better capture object contour and thin structures.
}
\label{fig:depth}
\end{figure*}

\subsection{Implementation details}
We implement our approach in TensorFlow~\cite{abadi2016tensorflow} and conduct all the experiments on a single Tesla K80 GPU with 12GB memory.
%
We set $\lambda_s = 3.0$, $\lambda_f=0.2$, and $\lambda_c=0.2$.
%
For network training, we use the Adam optimizer~\cite{kingma2014adam} with $\beta_1=0.9$, $\beta_2=0.99$.
%
%
%
In the following, we provide more implementation details in network architecture, network pre-training, and the proposed unsupervised joint training.

\paragraph{Network architecture.} For the pose network, we adopt the architecture from Zhou~\etal~\cite{zhou2017unsupervised}.
For the depth network, we use the ResNet-50~\cite{he2016deep} as our feature backbone with ELU~\cite{clevert2015fast} activation functions.
For the flow network, we adopt the UnFlow-C structure~\cite{meister2017unflow} --- a variant of FlowNetC~\cite{DFIB15}.
As our network training is \emph{model-agnostic}, more advanced network architectures (e.g., pose~\cite{godard2018digging}, depth~\cite{li2018megadepth}, or flow~\cite{sun2018pwc}) can be used for further improving the performance.

\paragraph{Unsupervised depth pre-training.} 
We train the depth and pose networks with a mini-batch size of 6 image pairs whose size is $576 \times 160$, from KITTI raw dataset or CityScapes dataset for 100K iterations.
We use a learning rate is 2e-4.
%
Each iteration takes around 0.8s (forward and backprop) during training. 

\paragraph{Unsupervised flow pre-training.} 
Following Meister~\etal~\cite{meister2017unflow}, we train the flow network with a mini-batch size of 4 image pairs whose size is $1152\times 320$ from SYNTHIA dataset for 300K iterations.
We keep the initial learning rate as 1e-4 for the first 100K iterations and then reduce the learning rate by half after each 100K iterations.
Each iteration takes around 2.4s (forward and backprop). 

\paragraph{Unsupervised joint training.}
We jointly train the depth, pose, and flow networks with a mini-batch size of 4 image pairs from KITTI raw dataset for 100K iterations.
Input size for the depth and pose networks is $576\times 160$, while the input size for the flow network is $1152\times 320$.
We divide the initial learning rate by 2 for every 20K iterations.
%
Our depth network produces depth predictions at 4 spatial scales, while the flow network produces flow fields at 5 scales.
We enforce the cross-network consistency in the finest 4 scales.
Each iteration takes around 3.6s (forward and backprop) during training.

\paragraph{Image resolution of network inputs/outputs.} As the input size of the UnFlow-C network~\cite{meister2017unflow} must be divisible by 64, we resize input image pairs of the two KITTI flow datasets to $1280 \times 384$ using bilinear interpolation. 
We then resize the estimated optical flow and rescale the predicted flow vectors to match the original input size.
For depth estimation, we resize the input image to the same size of training input to predict the disparity first.
We then resize and rescale the predicted disparity to the original size and compute the inverse the obtain the final prediction.

\begin{table*}[t]
\centering
\caption{\tb{Single-view depth estimation results} on \textit{test split} of KITTI raw dataset~\cite{Geiger2013IJRR}.
The methods trained on KITTI raw dataset~\cite{Geiger2013IJRR} are denoted by K. Models with additional training data from CityScapes \cite{Cordts2016Cityscapes} are denoted by CS+K. 
(D) denotes depth supervision, \red{(B)} denotes stereo input pairs, \blue{(M)} denotes monocular video clips
.
%
%
The best and the second best performance in each block are highlighted as bold and underline.
%
}
\label{tbl:kitti_depth_small}
\resizebox{1\textwidth}{!}{
\begin{tabular}{lcccccccccc}

\toprule
&& \multicolumn{4}{c}{Error metric $\downarrow$} && \multicolumn{3}{c}{Accuracy metric $\uparrow$} &
\\
\cmidrule{3-6} \cmidrule{8-10} 
Method  & Dataset & Abs Rel  & Sq Rel  & RMSE  & log RMSE  && $\delta < 1.25$ & $\delta < 1.25^2$ & $\delta < 1.25^3$\\

\midrule
Eigen~\etal~\cite{eigen2014depth} & K (D)  & 0.203 & 1.548 & 6.307 & 0.246 && 0.702 & 0.890 & 0.958 \\
Kuznietsov~\etal~\cite{kuznietsov2017semi} & K \red{(B)} / K (D) & \textbf{0.113} & \textbf{0.741} & \textbf{4.621} & \textbf{0.189} && \textbf{0.862} & \textbf{0.960} & \textbf{0.986} \\
Zhan~\etal~\cite{zhan2018unsupervised} &  K \red{(B)} & 0.144 & 1.391 & 5.869 &  0.241 && 0.803 & 0.928 & 0.969 \\
Godard~\etal~\cite{godard2016unsupervised} & K \red{(B)} & 0.133 & 1.140 & 5.527 & 0.229  && 0.830 & 0.936 & 0.970 \\
Godard~\etal~\cite{godard2016unsupervised} & CS+K \red{(B)} & \underline{0.121} & \underline{1.032} & \underline{5.200} & \underline{0.215}  && \underline{0.854} & \underline{0.944} & \underline{0.973}\\

\midrule
Zhou~\etal~\cite{zhou2017unsupervised} & K \blue{(M)} & 0.208 & 1.768 & 6.856 & 0.283 && 0.678 & 0.885 & 0.957 \\
Yang~\etal~\cite{yang2017unsupervised} & K \blue{(M)} & 0.182 & 1.481 & 6.501 & 0.267 && 0.725 & 0.906 & 0.963 \\
Mahjourian~\etal~\cite{mahjourian2018unsupervised} & K \blue{(M)} & 0.163 & 1.240 & 6.220 & 0.250 && 0.762 & 0.916 & 0.968 \\
Yang~\etal~\cite{yang2018lego} & K \blue{(M)} & 0.162 & 1.352 & 6.276 & 0.252 && - & - & - \\
Yin~\etal~\cite{yin2018geonet} & K \blue{(M)} & 0.155 & 1.296 & 5.857 & 0.233 && 0.793 & 0.931 & \textbf{0.973} \\
Godard~\etal~\cite{godard2018digging} & K \blue{(M)} & \underline{0.154} & \underline{1.218} & 5.699 & 0.231 && 0.798 & \underline{0.932} & \textbf{0.973}  \\

Ours (w/o forward-backward) & K \blue{(M)} & 0.160 & 1.256 & 5.555 & 0.226 && 0.796 & 0.931 & \textbf{0.973} \\
Ours (w/o cross-task) & K \blue{(M)} & 0.160 & 1.234 & \underline{5.508} & \underline{0.225} && \underline{0.800} & \underline{0.932} & \underline{0.972} \\
Ours & K \blue{(M)} & \textbf{0.150} & \textbf{1.124} & \textbf{5.507} & \textbf{0.223} && \textbf{0.806} & \textbf{0.933} & \textbf{0.973} \\

\midrule
Zhou~\etal~\cite{zhou2017unsupervised} & CS+K \blue{(M)} & 0.198 & 1.836 & 6.565 & 0.275 && 0.718 & 0.901 & 0.960\\
Yang~\etal~\cite{yang2017unsupervised} & CS+K \blue{(M)} & 0.165 & 1.360 & 6.641 & 0.248 && 0.750 & 0.914 & 0.969 \\
Mahjourian~\etal~\cite{mahjourian2018unsupervised} & CS+K \blue{(M)} & 0.159 & 1.231 & 5.912 & 0.243 && 0.784 & 0.923 & 0.970 \\
Yang~\etal~\cite{yang2018lego} & CS+K \blue{(M)} & 0.159 & 1.345 & 6.254 & 0.247 && - & - & - \\
Yin~\etal~\cite{yin2018geonet} & CS+K \blue{(M)} & \underline{0.153} & 1.328 & 5.737 & 0.232 && 0.802 & 0.934 & 0.972 \\

Ours (w/o forward-backward) & CS+K \blue{(M)} & 
0.159 & 1.716 & 5.616 & 0.222 && \underline{0.805} & \underline{0.939} & 0.976 \\
Ours (w/o cross-task) & CS+K \blue{(M)} & 0.155 & \textbf{1.181} & \underline{5.301} & \underline{0.218} && \underline{0.805} & \underline{0.939} & \underline{0.977} \\
Ours & CS+K \blue{(M)} & \textbf{0.146} & \underline{1.182} & \textbf{5.215} & \textbf{0.213} && \textbf{0.818} & \textbf{0.943} & \textbf{0.978} \\


\bottomrule
\end{tabular}
}
\label{tbl:kitti_depth}
\end{table*}

\subsection{Evaluation metrics}
Following Zhou~\etal~\cite{zhou2017unsupervised}, we evaluate our depth network using several error metrics (absolute relative difference, square related difference, RMSE, log RMSE).
For optical flow estimation, we compute the average endpoint error (EPE) on pixels with the ground truth flow available for each dataset. 
On KITTI flow 2015 dataset~\cite{Menze2015CVPR}, we also compute the F1 score, which is the percentage of pixels that have EPE greater than 3 pixels and 5\% of the ground truth value.

%

\subsection{Experimental evaluation}
\paragraph{Single-view depth estimation.}
We compare our depth network with state-of-the-art algorithms on the \emph{test split} of the KITTI raw dataset provided by Eigen~\etal~\cite{eigen2014depth}.
As shown in Table~\ref{tbl:kitti_depth_small}, our method 
achieves the state-of-the-art performance when compared with models trained with monocular video sequences.
However, our method performs slightly worse than the models that exploit calibrated stereo image pairs (\ie pose supervision) or with additional ground truth depth annotation.
We believe that performance gap can be attributed to the error induced by our pose network.
Extending our approach to \emph{calibrated stereo videos} is an interesting future direction.

We also conduct an ablation study by removing the forward-backward consistency loss or cross-task consistency loss. 
In both cases our results show significant performance of degradation, highlighting the importance the proposed consistency loss.
%
\figref{depth} shows qualitative comparison with \cite{eigen2014depth,zhou2017unsupervised}, our method can better capture thin structure and delineate clear object contour.

To evaluate the generalization ability of our depth network on general scenes, we also apply our trained model to the Make3D dataset~\cite{saxena2006learning}.
Table~\ref{tbl:make3d_depth} shows that our method achieves the state-of-the-art performance compared with existing unsupervised models and is competitive with respect to supervised learning models (even without fine-tuning on Make3D datasets).
%

\begin{table}[htbp]
\centering
\caption{\tb{Results on the Make3D dataset~\cite{saxena20083}}. 
Our results were obtained by the model trained on Cityscapes + KITTI \emph{without} fine-tuning on the training images in Make3D. 
Following the evaluation protocol of ~\cite{godard2016unsupervised}, the errors are only computed where depth is less than 70 meters. The best and the second best performance in each block are highlighted as bold and underline.
}


\begin{tabular}{lccccc}
\toprule
&& \multicolumn{4}{c}{Error metric $\downarrow$}
\\
\cmidrule{3-6}
Method & Supervision & Abs Rel & Sq Rel & RMSE & log RMSE \\

\midrule
Train set mean &-& 0.876 & 12.98 & 12.27 & 0.307 \\
Karsch~\etal~\cite{karsch2014depth} &depth& {0.428} & \underline{5.079} & {8.389} & {0.149} \\
Liu~\etal~\cite{liu2014discrete} &depth& 0.475 & 6.562 & 10.05 & 0.165 \\
Laina~\etal~\cite{laina2016deeper} &depth& \underline{0.204} & \textbf{1.840} & \underline{5.683} & \underline{0.084} \\
Li~\etal~\cite{li2018megadepth} &depth&
\textbf{0.176} & - & \textbf{4.260} & \textbf{0.069} \\
\midrule
Godard~\etal~\cite{godard2016unsupervised} &pose& 0.544 & 10.94 & 11.76 & \textbf{0.193} \\
Zhou~\etal~\cite{zhou2017unsupervised} & none & \underline{0.383} & \underline{5.321} & \underline{10.47} & 0.478 \\

Ours & none & \textbf{0.331} & \textbf{2.698} & \textbf{6.89} & \underline{0.416} \\

\bottomrule

\end{tabular}
\label{tbl:make3d_depth}

\end{table}

\begin{table}[htbp]
\centering
\caption{\tb{Quantitative evaluation on optical flow.} Results on KITTI flow 2012 ~\cite{Geiger2012CVPR} , KITTI flow 2015~\cite{Menze2015CVPR} datasets. 
We denote ``C'' as the FlyingChairs dataset~\cite{DFIB15}, ``T'' as the FlyingThings3D dataset~\cite{mayer2016large}, ``K'' as the KITTI raw dataset~\cite{Geiger2013IJRR}, ``SYN'' as the SYNTHIA dataset~\cite{Ros_2016_CVPR}.
%
\red{(S)} indicates that the model is trained with ground truth annotation, while \blue{(U)} indicates the model is trained in an unsupervised manner.
The best and the second best performance in each block are highlighted as bold and underline.
}

\resizebox{1\textwidth}{!}{

\begin{tabular}{lccccccccccccc}
\toprule
&& \multicolumn{2}{c}{KITTI 2012} && 
\multicolumn{3}{c}{KITTI 2015}
\\
\cmidrule{3-4} \cmidrule{6-8}
&& Train & Test && Train & Train & Test\\
Method  & Dataset & EPE & EPE && EPE & F1 & F1\\

\midrule
LDOF~\cite{brox2009large} &-& 10.94 & 12.4 && 18.19 & 38.05\% & - \\

DeepFlow~\cite{weinzaepfel2013deepflow} &-& 4.58 & \underline{5.8} && 10.63 & \underline{26.52\%} & \underline{29.18\%} \\

EpicFlow~\cite{revaud2015epicflow} &-& \underline{3.47} & \textbf{3.8} && \underline{9.27} & 27.18\% & \textbf{27.10\%} \\

FlowField~\cite{bailer2015flow} &-& \textbf{3.33} & - && \textbf{8.33} & \textbf{24.43\%} & - \\

\midrule
FlowNetS~\cite{DFIB15} & C \red{(S)} & 8.26 & - && 15.44 & 52.86\% & - \\

FlowNetC~\cite{DFIB15} & C \red{(S)} & 9.35 &- && \underline{12.52} & 47.93\% & -  \\

SpyNet~\cite{ranjan2016optical} & C \red{(S)} & 9.12 &-&& 20.56 & 44.78\% & - \\

SemiFlowGAN~\cite{Lai-NIPS-2017} &C \red{(S)} / K \blue{(U)}& \underline{7.16} &-&& 16.02 & \underline{38.77\%} &- \\

FlowNet2~\cite{ilg2016flownet} &C \red{(S)} + T \red{(S)}& \textbf{4.09} &-&& \textbf{10.06} & \textbf{30.37\%} &- \\

\midrule
UnsupFlownet~\cite{jason2016back} & C \blue{(U)} + K \blue{(U)}& 11.3 & 9.9 && - & - &- \\

DSTFlow~\cite{ren2017unsupervised} & C \blue{(U)} & 16.98 & - && 24.30 & 52.00\% & - \\

DSTFlow~\cite{ren2017unsupervised} & K \blue{(U)} & 10.43 & 12.4 && 16.79 & 36.00\% & 39.00\% \\

Yin~\etal~\cite{yin2018geonet} & K \blue{(U)} & - & - && 10.81 & - & - \\


UnFlowC~\cite{meister2017unflow} & SYN \blue{(U)} + K \blue{(U)}  & \underline{3.78} & \underline{4.5} && \textbf{8.80} & 28.94\% & 29.46\% \\

Ours (w/o forward-backward) & SYN \blue{(U)} + K \blue{(U)} & 3.86 & 4.7 && 9.12 & \underline{26.27\%} & \underline{26.90\%} \\

Ours (w/o cross-task) & SYN \blue{(U)} + K \blue{(U)} & 4.70 & 5.8 && \underline{8.95} & 28.37\% & 30.03\% \\

{Ours} & SYN \blue{(U)} + K \blue{(U)} & \textbf{3.54} & \textbf{4.4} && 8.98 & \textbf{26.01\%} & \textbf{25.70\%} \\

\midrule
FlowNet2-ft-kitti~\cite{ilg2016flownet} & C \red{(S)} + T \red{(S)} + K \red{(S)} & \underline{(1.28)} & \underline{1.8} && \underline{(2.30)} & \underline{(8.61\%)} & \underline{11.48\%} \\

UnFlowCSS-ft-kitti~\cite{meister2017unflow} & SYN \blue{(U)} + K \blue{(U)} + K \red{(S)}& \textbf{(1.14)} & \textbf{1.7} && \textbf{(1.86)} & \textbf{(7.40\%)} & \textbf{11.11\%} \\

UnFlowC-ft-kitti~\cite{meister2017unflow} & SYN \blue{(U)} + K \blue{(U)} + K \red{(S)}& (2.13) & 3.0 && (3.67) & (17.78\%) & 24.20\% \\

{Ours-ft-kitti} & SYN \blue{(U)} + K \blue{(U)} + K \red{(S)}& (1.75) & 3.0 && (2.85) & (13.47\%) & 22.82\% \\

\bottomrule
\end{tabular}

}

\label{tbl:kitti_flow}
\end{table}

\begin{figure*}[t]
\centering

\mpage{0.15}{\includegraphics[width=1.0\linewidth]{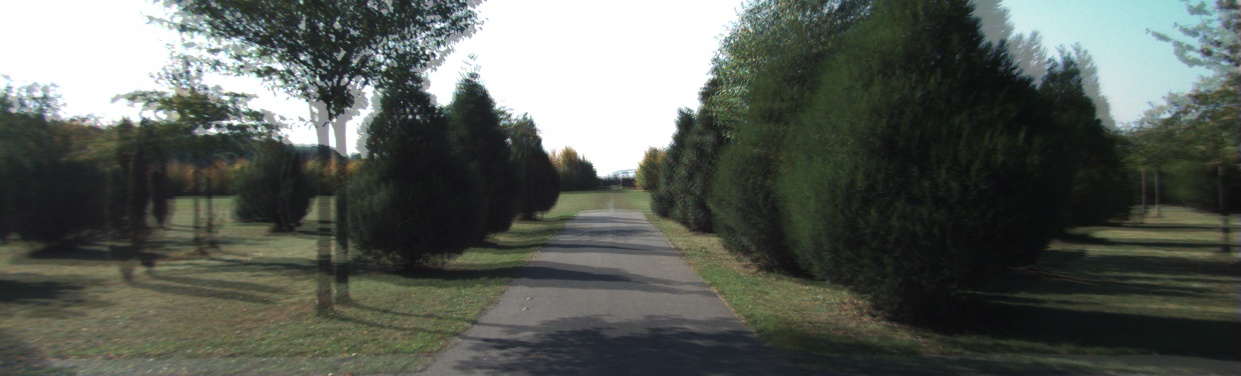}}\hfill
\mpage{0.15}{\includegraphics[width=1.0\linewidth]{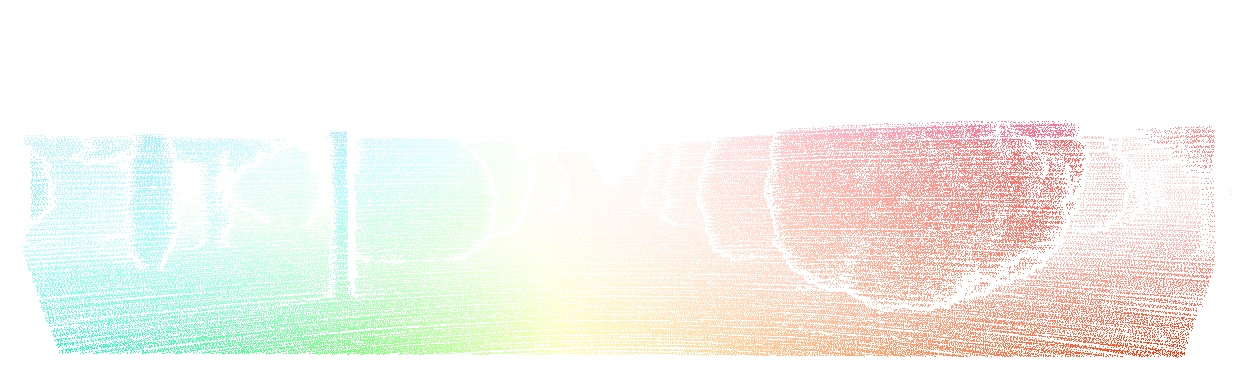}}\hfill
\mpage{0.15}{\includegraphics[width=1.0\linewidth]{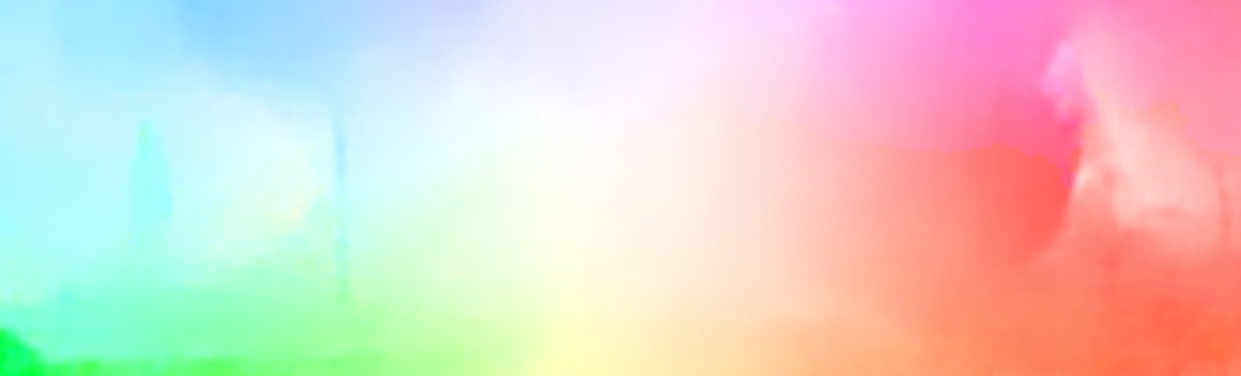}}\hfill
\mpage{0.15}{\includegraphics[width=1.0\linewidth]{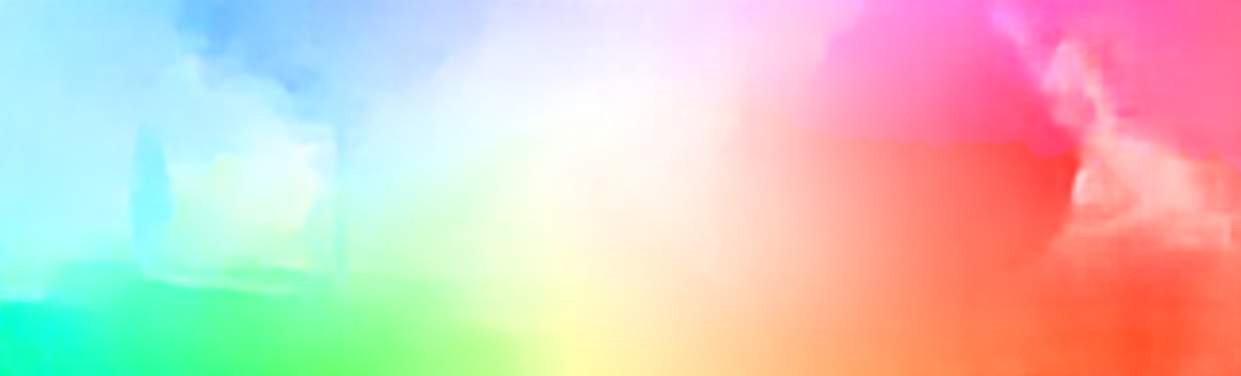}}\hfill
\mpage{0.15}{\includegraphics[width=1.0\linewidth]{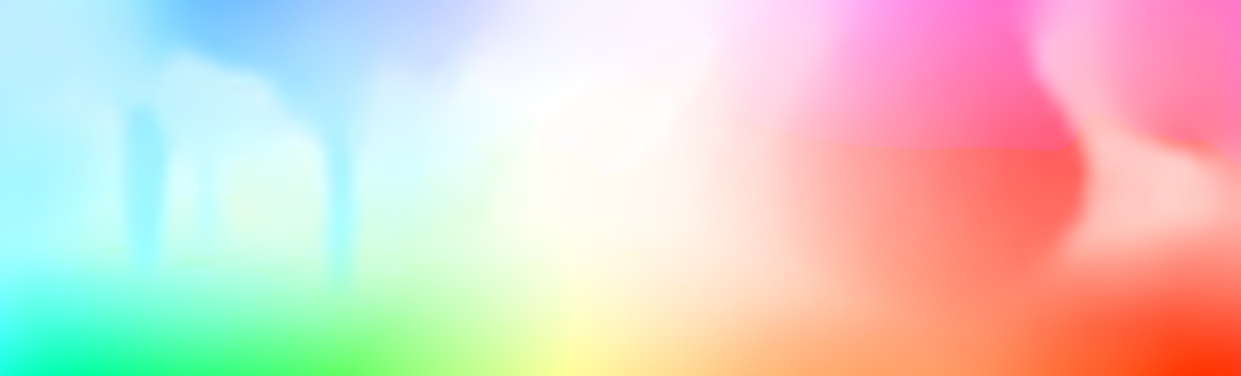}}\hfill
\mpage{0.15}{\includegraphics[width=1.0\linewidth]{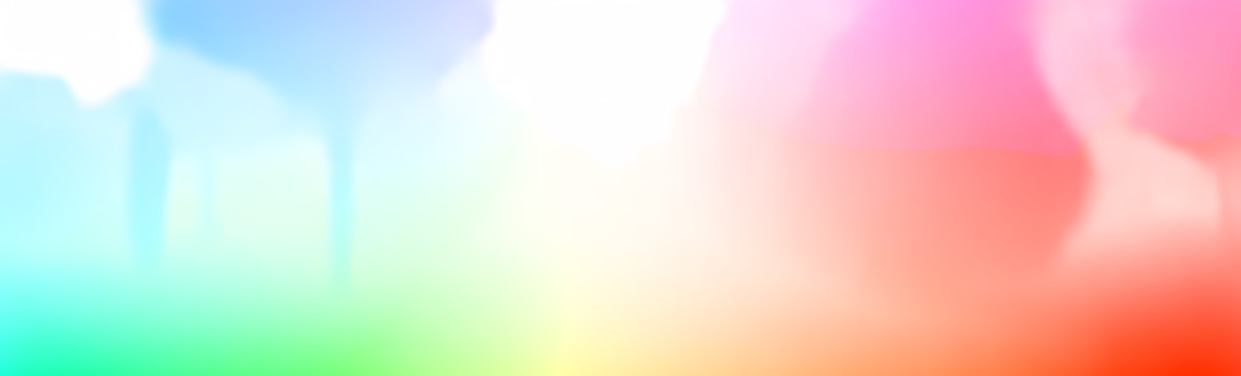}}\hfill




\mpage{0.15}{\includegraphics[width=1.0\linewidth]{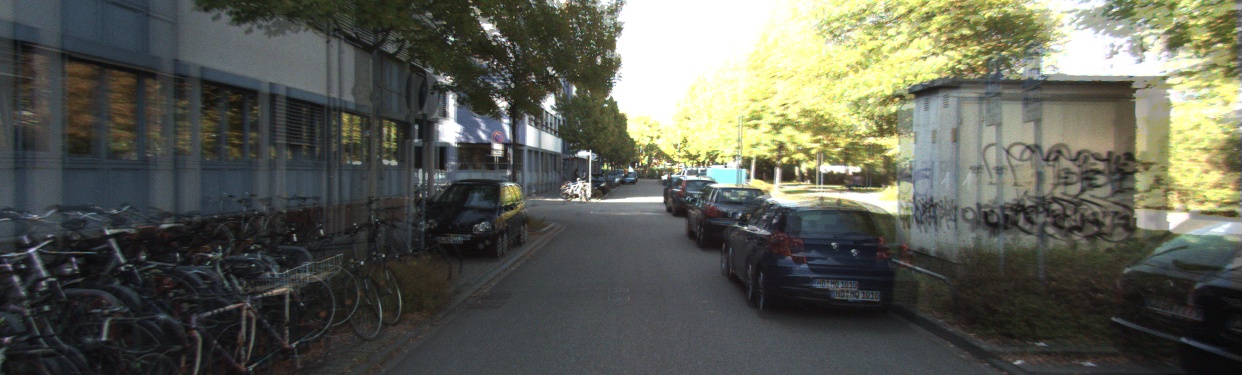}}\hfill
\mpage{0.15}{\includegraphics[width=1.0\linewidth]{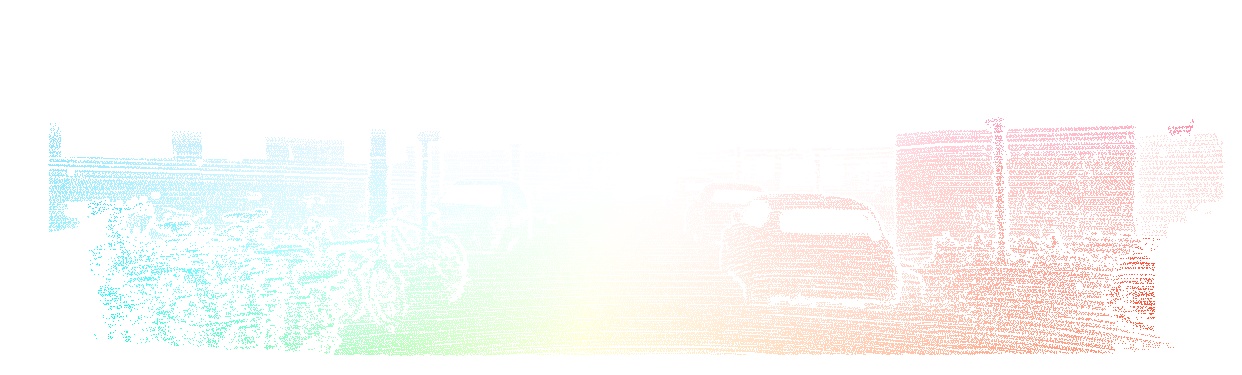}}\hfill
\mpage{0.15}{\includegraphics[width=1.0\linewidth]{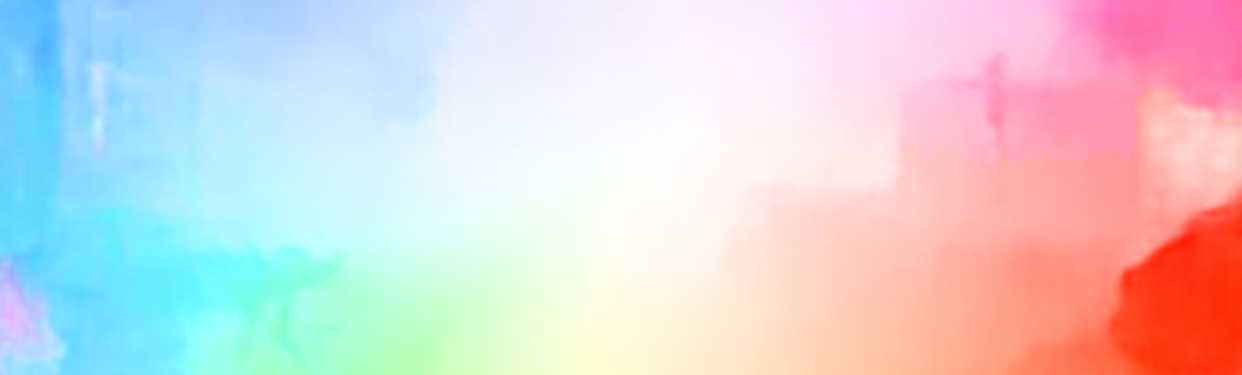}}\hfill
\mpage{0.15}{\includegraphics[width=1.0\linewidth]{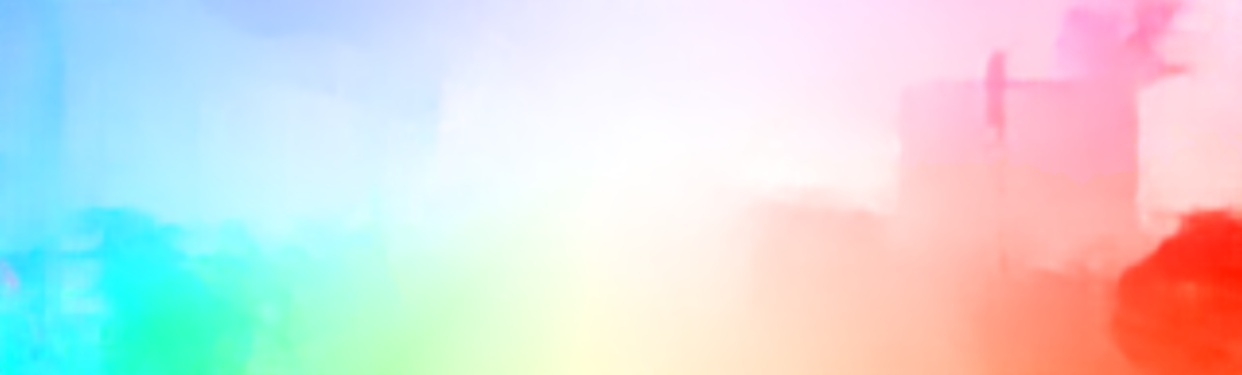}}\hfill
\mpage{0.15}{\includegraphics[width=1.0\linewidth]{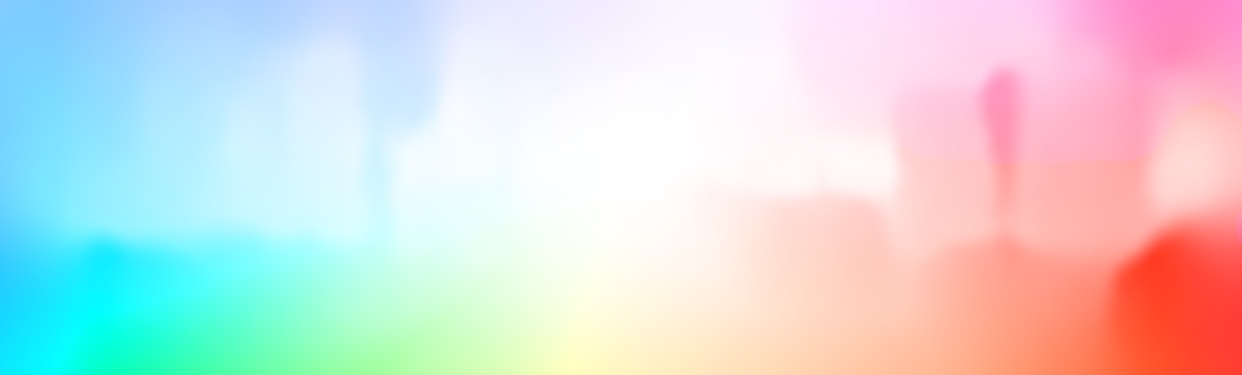}}\hfill
\mpage{0.15}{\includegraphics[width=1.0\linewidth]{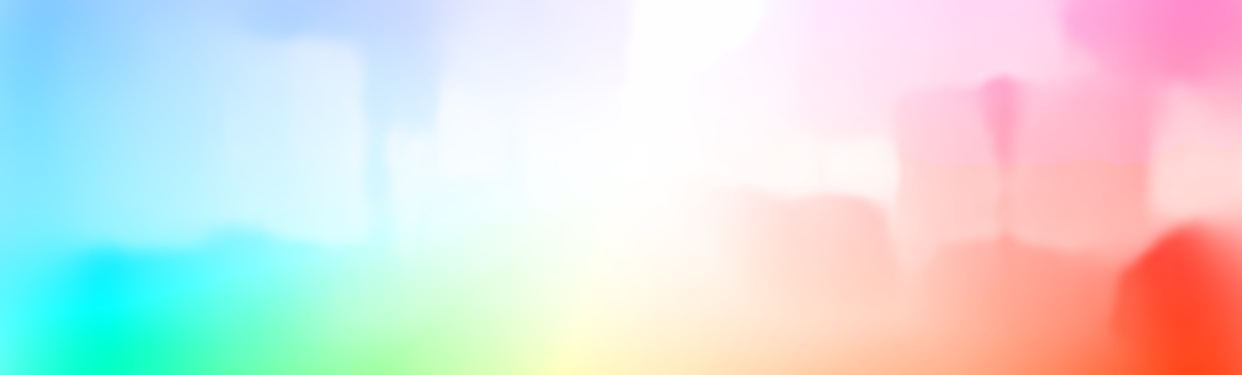}}\hfill


\mpage{0.15}{\includegraphics[width=1.0\linewidth]{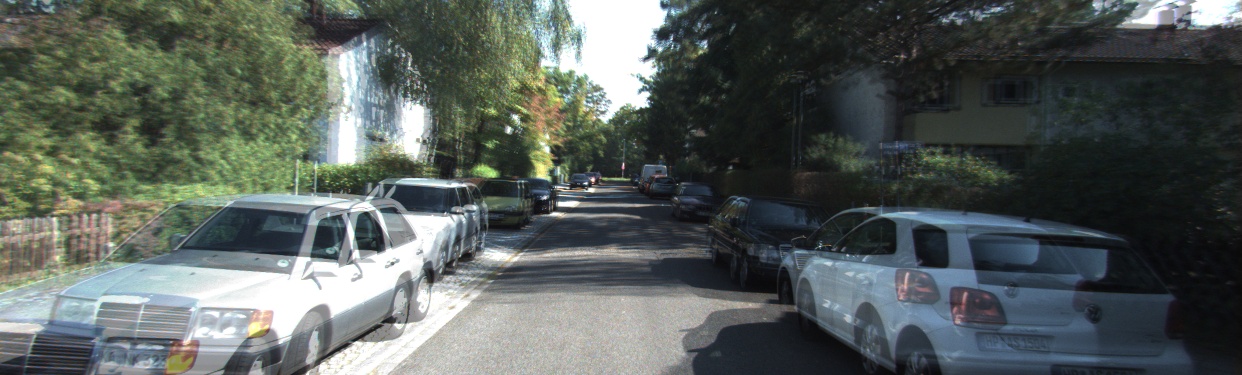}}\hfill
\mpage{0.15}{\includegraphics[width=1.0\linewidth]{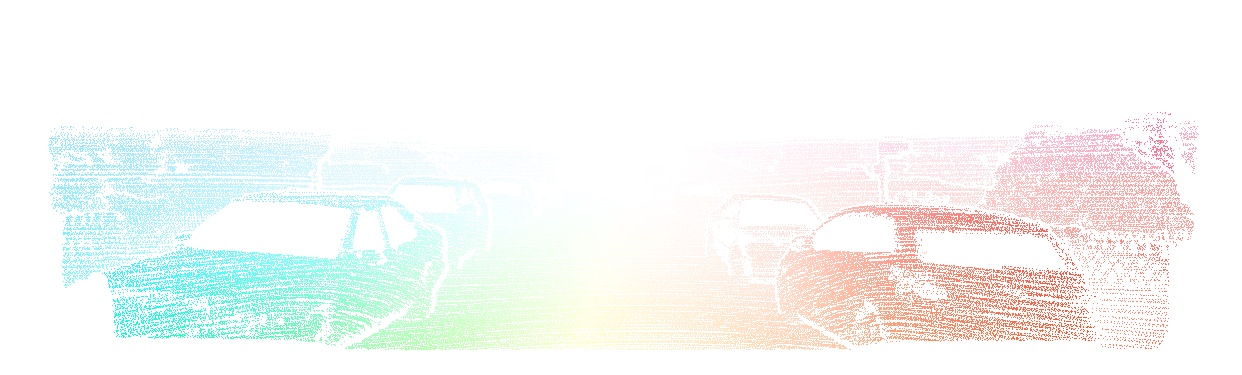}}\hfill
\mpage{0.15}{\includegraphics[width=1.0\linewidth]{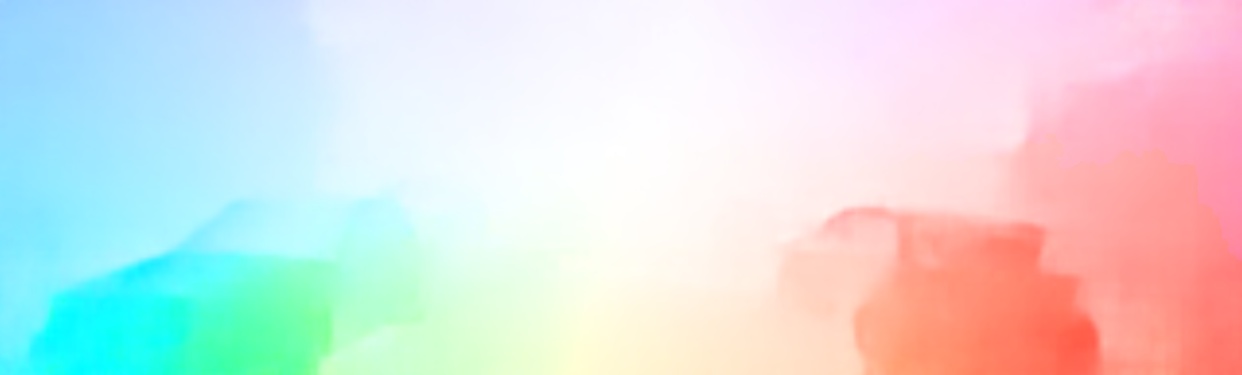}}\hfill
\mpage{0.15}{\includegraphics[width=1.0\linewidth]{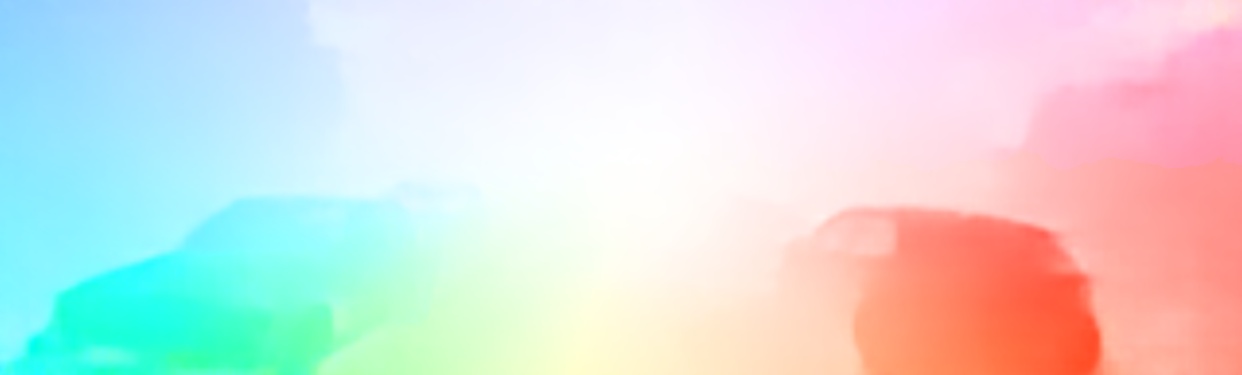}}\hfill
\mpage{0.15}{\includegraphics[width=1.0\linewidth]{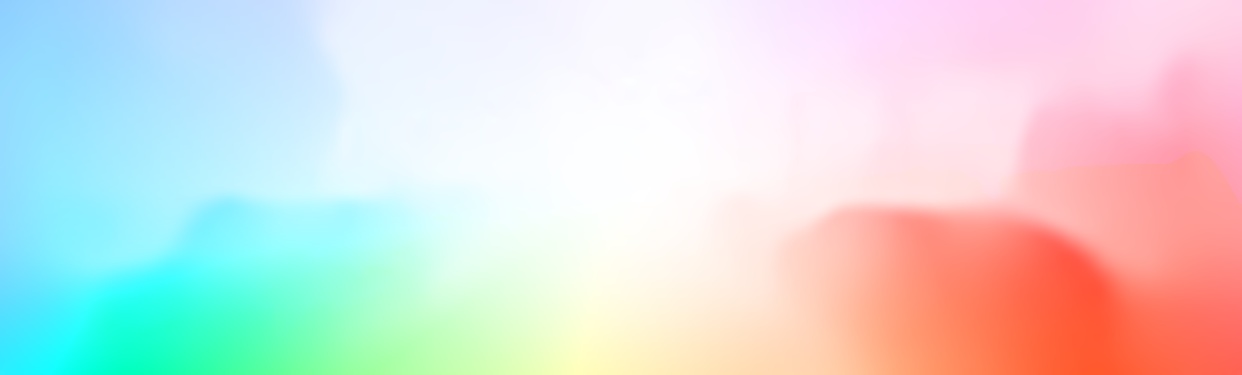}}\hfill
\mpage{0.15}{\includegraphics[width=1.0\linewidth]{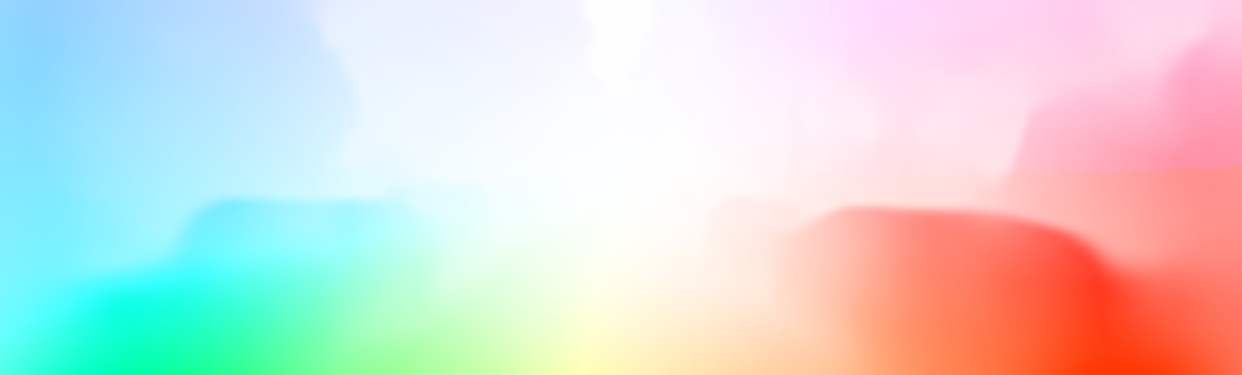}}\hfill

\mpage{0.15}{\includegraphics[width=1.0\linewidth]{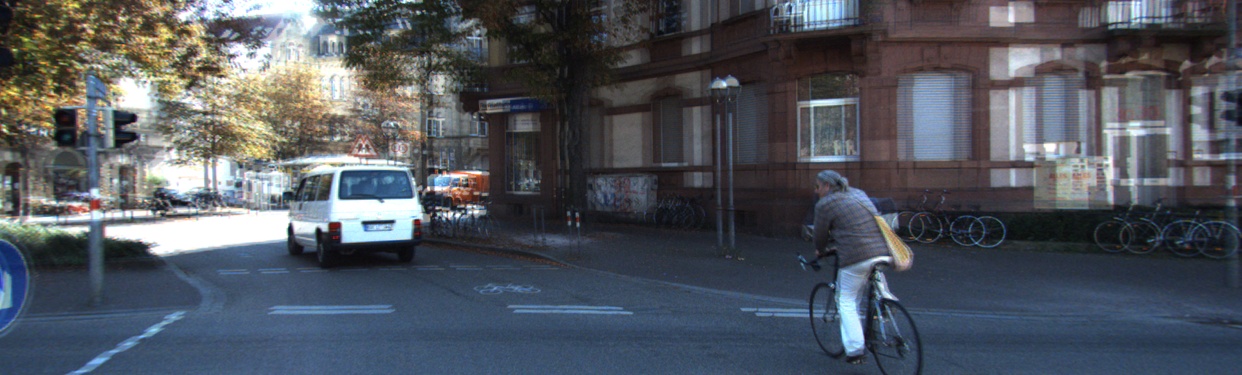}}\hfill
\mpage{0.15}{\includegraphics[width=1.0\linewidth]{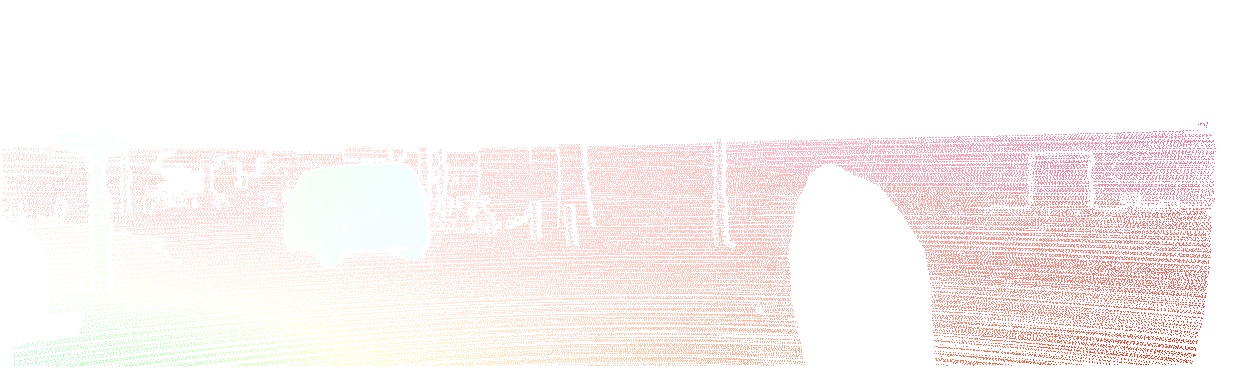}}\hfill
\mpage{0.15}{\includegraphics[width=1.0\linewidth]{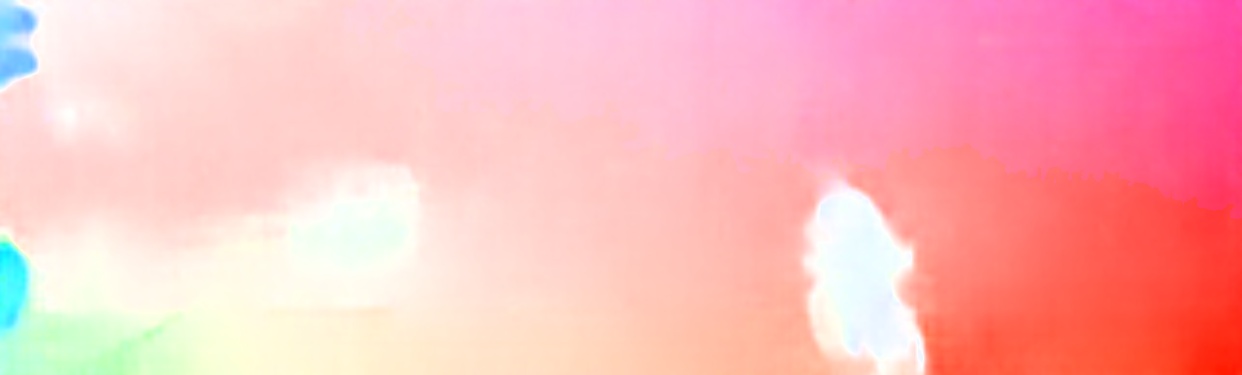}}\hfill
\mpage{0.15}{\includegraphics[width=1.0\linewidth]{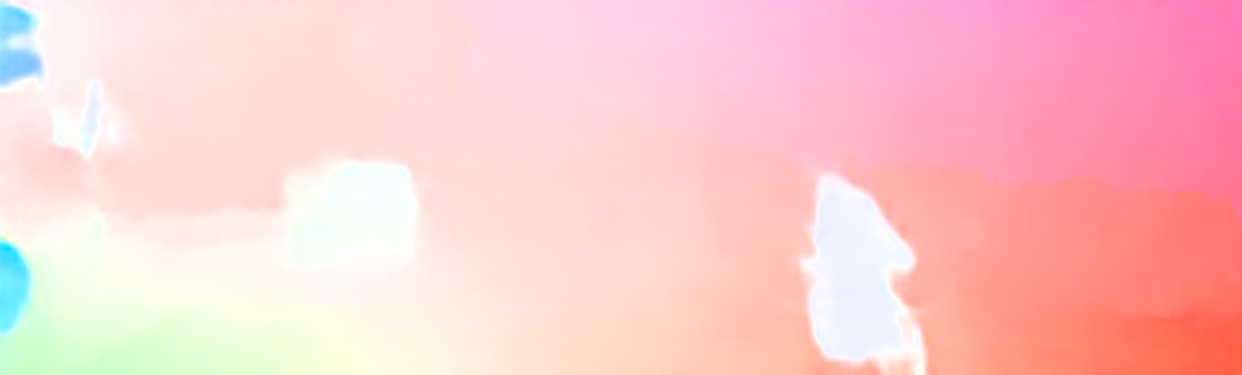}}\hfill
\mpage{0.15}{\includegraphics[width=1.0\linewidth]{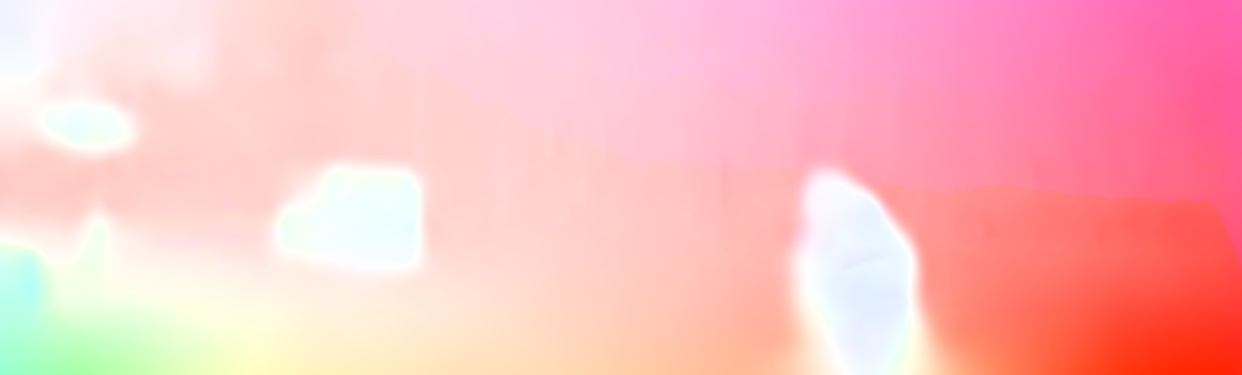}}\hfill
\mpage{0.15}{\includegraphics[width=1.0\linewidth]{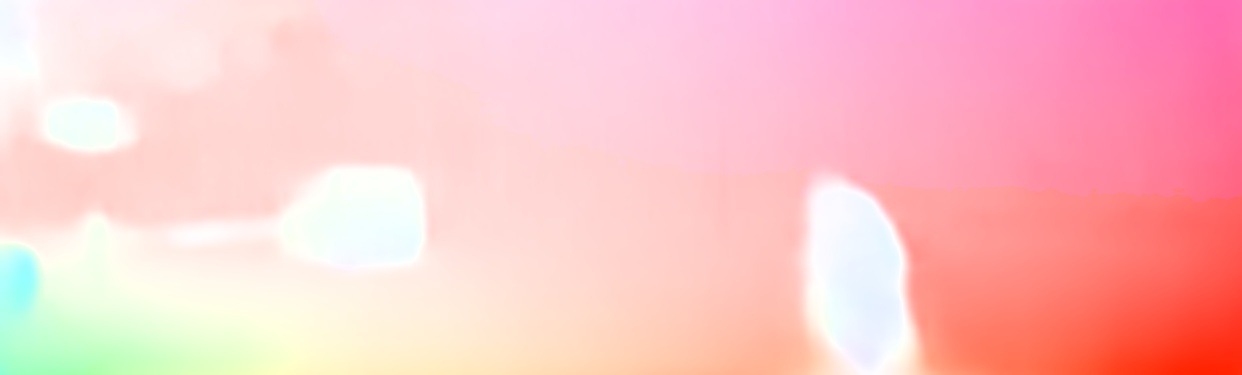}}\hfill


\mpage{0.15}{\includegraphics[width=1.0\linewidth]{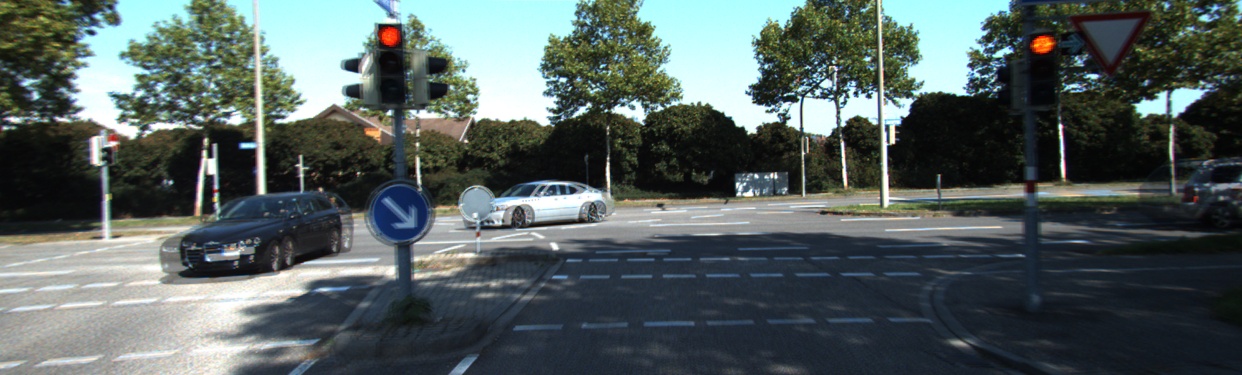}}\hfill
\mpage{0.15}{\includegraphics[width=1.0\linewidth]{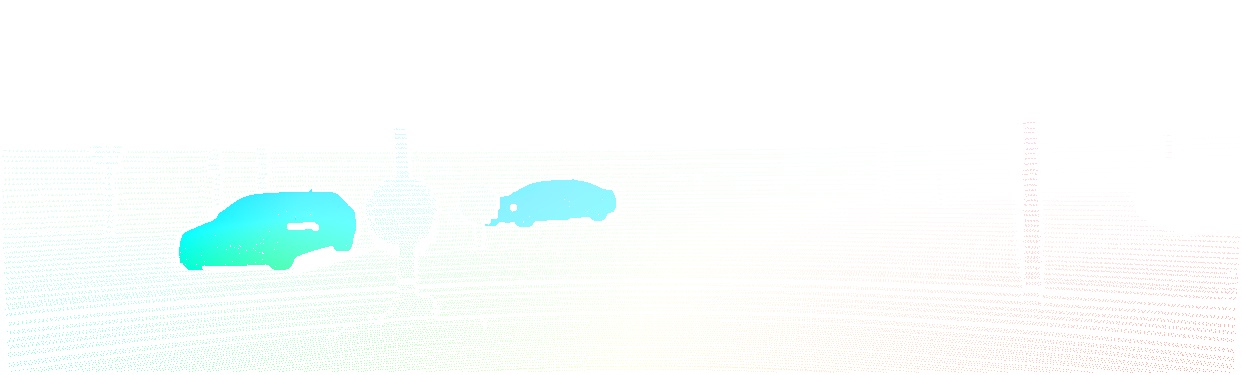}}\hfill
\mpage{0.15}{\includegraphics[width=1.0\linewidth]{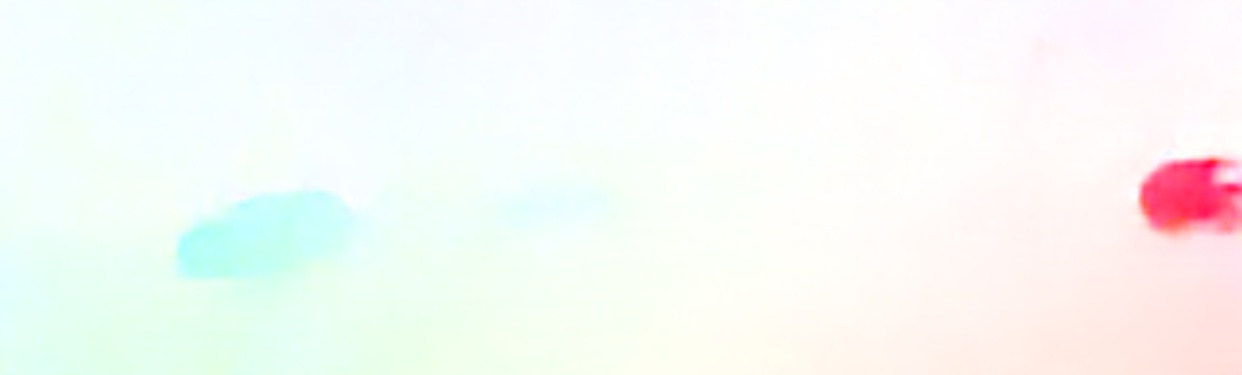}}\hfill
\mpage{0.15}{\includegraphics[width=1.0\linewidth]{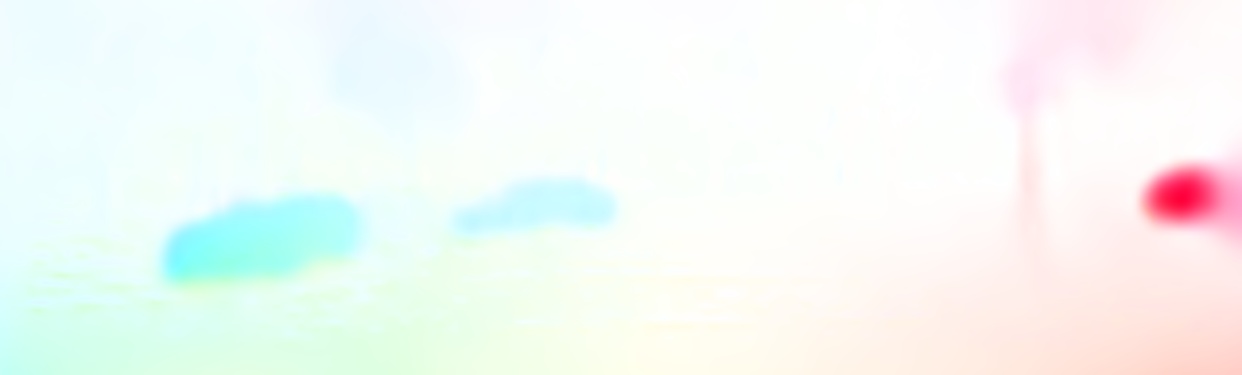}}\hfill
\mpage{0.15}{\includegraphics[width=1.0\linewidth]{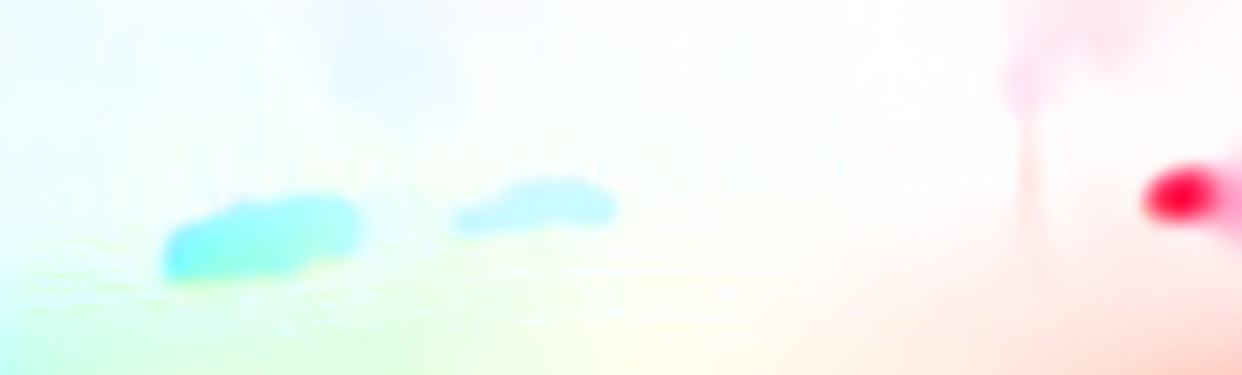}}\hfill
\mpage{0.15}{\includegraphics[width=1.0\linewidth]{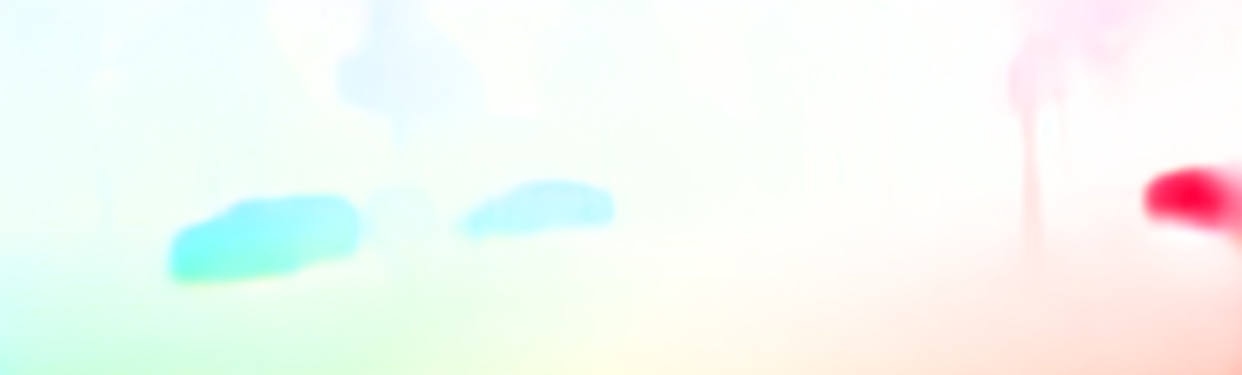}}\hfill


\mpage{0.15}{\includegraphics[width=1.0\linewidth]{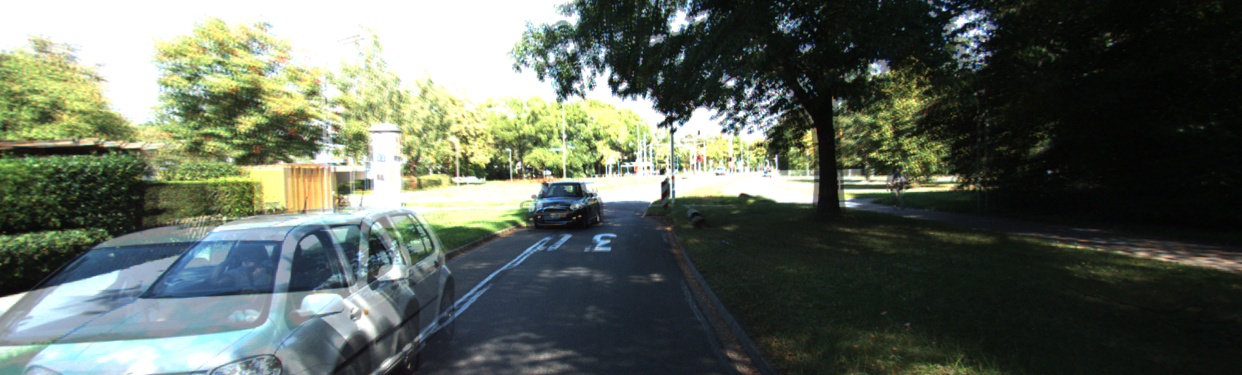}}\hfill
\mpage{0.15}{\includegraphics[width=1.0\linewidth]{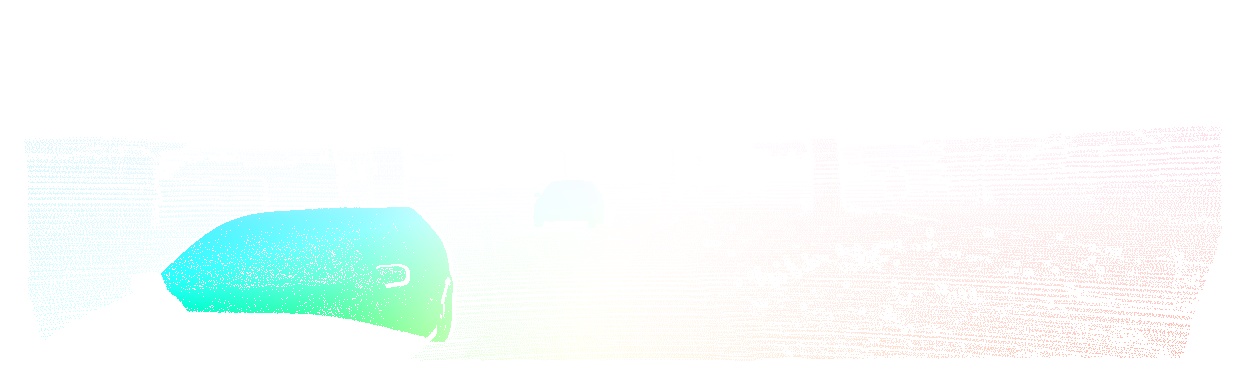}}\hfill
\mpage{0.15}{\includegraphics[width=1.0\linewidth]{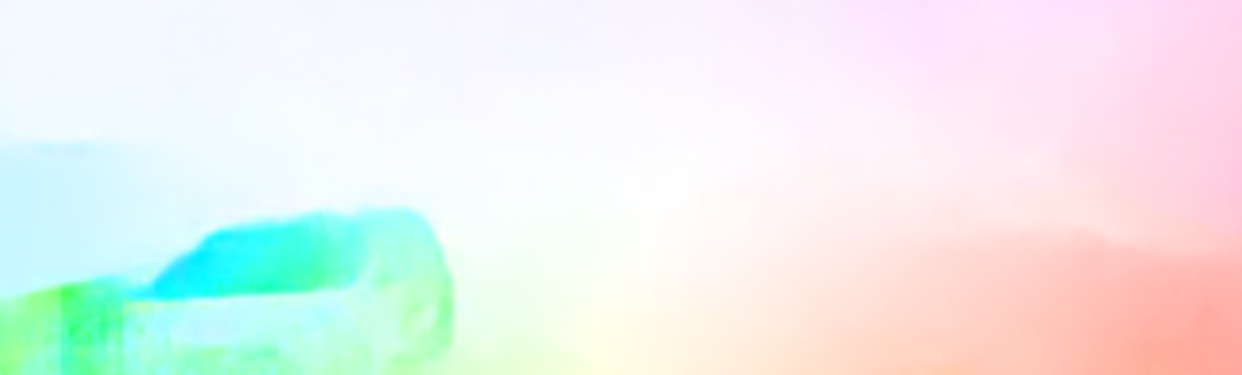}}\hfill
\mpage{0.15}{\includegraphics[width=1.0\linewidth]{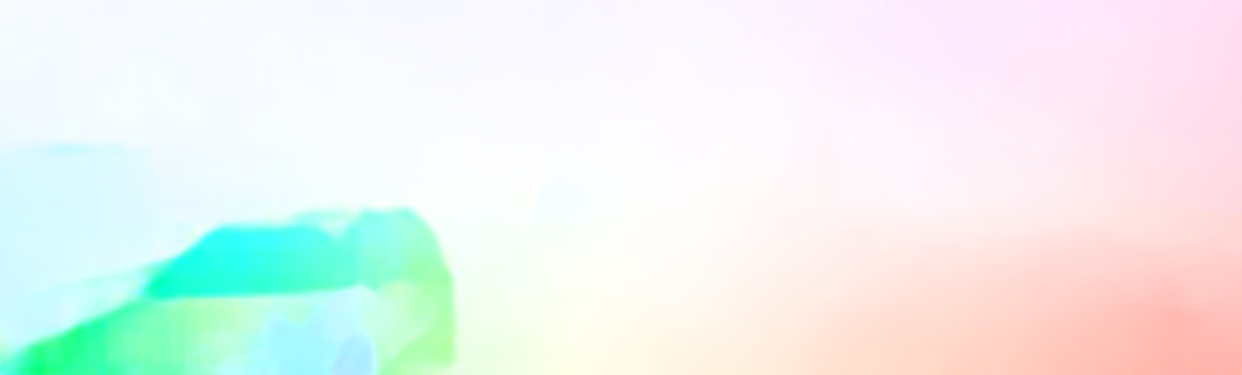}}\hfill
\mpage{0.15}{\includegraphics[width=1.0\linewidth]{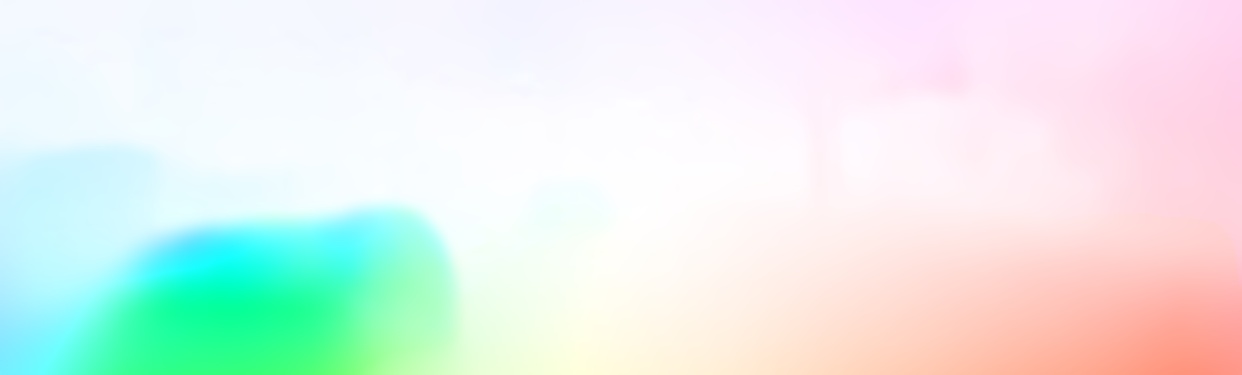}}\hfill
\mpage{0.15}{\includegraphics[width=1.0\linewidth]{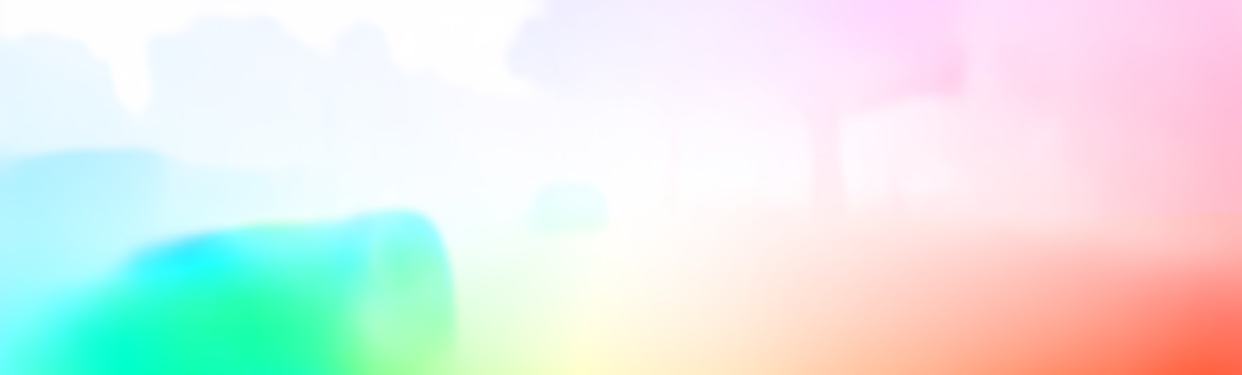}}\hfill

\mpage{0.12}{Input} \hfill
\mpage{0.18}{Ground truth} \hfill
\mpage{0.13}{FlowNetS} \hfill
\mpage{0.13}{FlowNetC} \hfill
\mpage{0.13}{UnFlow-C} \hfill
\mpage{0.13}{Ours} \hfill

\caption{\textbf{Visual results on KITTI flow datasets.}
All the models are directly applied \emph{without} fine-tuning on KITTI flow annotations.
Our model delineates clearer object contours compared to both supervised/unsupervised methods.
}
\label{fig:flow}
\end{figure*}
\begin{table}[htbp]
\centering

\caption{\textbf{Pose estimation results} on KITTI Odometry datest~\cite{Geiger2012CVPR}.}

\begin{tabular}{lccc}
\toprule
 & Seq. 09 && Seq. 10\\
\midrule
ORB-SLAM (full) & 0.014$\pm$0.008 && 0.012$\pm$0.011 \\
\midrule
ORB-SLAM (short) & 0.064$\pm$0.141 && 0.064$\pm$0.130 \\
Mean Odom. & 0.032$\pm$0.026 && 0.028$\pm$0.023 \\
Zhou~\etal~\cite{zhou2017unsupervised} & 0.021$\pm$0.017 && 0.020$\pm$0.015 \\
Mahjourian~\etal~\cite{mahjourian2018unsupervised} & 0.013$\pm$0.010 && 0.012$\pm$0.011 \\
Yin~\etal~\cite{yin2018geonet} & \textbf{0.012$\pm$0.007} && 
\textbf{0.012$\pm$0.009} \\
Ours & 0.017$\pm$0.007 && 0.015$\pm$0.009 \\
\bottomrule
\end{tabular}
\label{tbl:kitti_pose}
\end{table}

\paragraph{Optical flow estimation.}
We compare our flow network with 
conventional variational algorithms, supervised CNN methods, and several unsupervised CNN models
on the KITTI flow 2012 and 2015 datasets.
As shown in Table~\ref{tbl:kitti_flow}, our method achieves state-of-the-art performance on both datasets.
A visual comparison can be found in~\figref{flow}.
With optional fine-tuning on available ground truth labels on the KITTI flow datasets, we show that our approach achieves competitive performance sharing similar network architectures.
%
This suggests that our method can serve as an unsupervised pre-training technique for learning optical flow in domains where the amounts of ground truth data are scarce. 

\paragraph{Pose estimation.}
For completeness, we provide the performance evaluation of the pose network. 
We follow the same evaluation protocol as \cite{zhou2017unsupervised} and use a 5-frame based pose network.
As shown in Table~\ref{tbl:kitti_pose}, our pose network shows competitive performance with respect to state-of-the-art visual SLAM methods or other unsupervised learning methods.
We believe that a better pose network would further improve the performance of both depth or optical flow estimation.

\section{Conclusions}
\label{sec:conclusions}

We presented an unsupervised learning framework for both sing-view depth prediction and optical flow estimation using unlabeled video sequences.
Our key technical contribution lies in the proposed cross-task consistency that couples the network training.
At test time, the trained depth and flow models can be applied independently. 
We validate the benefits of joint training through extensive experiments on benchmark datasets.
Our single-view depth prediction model compares favorably against existing unsupervised models using unstructured videos on both KITTI and Make3D datasets.
Our flow estimation model achieves competitive performance with state-of-the-art approaches. 
By leveraging geometric constraints, our work suggests a promising future direction of advancing the state-of-the-art in multiple dense prediction tasks using unlabeled data.

\paragraph{Acknowledgement.} This work was supported in part by NSF under Grant No. (\#1755785). We thank NVIDIA Corporation for the donation of GPUs.

\clearpage
{\small
\bibliographystyle{splncs04}
\bibliography{main}
}


\end{document}